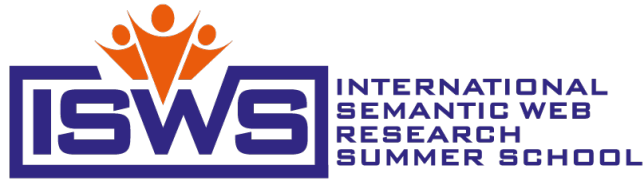

# Semantic Web and Creative AI

A Technical Report from ISWS 2023
June 11th - 17th, 2023

January 30, 2025

Bertinoro, Italy

# Authors

**Editors**
Stefano De Giorgis, National Research Council (CNR), Italy
Tabea Tietz, FIZ Karlsruhe and Karlsruhe Insitute of Technology, Germany

**Directors**
Valentina Presutti, Semantic Technology Lab, ISTC-CNR, Rome, Italy
Harald Sack, FIZ Karlsruhe and Karlsruhe Insitute of Technology, Germany

**Tutors**
Claudia d'Amato, University of Bari, Italy
Irene Celino, Cefriel, Milan, Italy
John Domingue, KMi, Open University and President of STI International, United Kingdom
Michel Dumontier, Maastricht University, Netherlands
Frank van Harmelen, Vrije Universiteit Amsterdam, Netherlands
Aidan Hogan, Universidad de Chile, Chile
Jan-Christoph Kalo, Vrije Universiteit, Netherlands
Valentina Presutti, Semantic Technology Lab, ISTC-CNR, Rome, Italy
Sebastian Rudolph, TU Dresden, Germany
Harald Sack, FIZ Karlsruhe and Karlsruhe Insitute of Technology, Germany

**Assistant Tutors**
Oleksandra Bruns, FIZ Karlsruhe and Karlsruhe Insitute of Technology, Germany
Stefano De Giorgis, National Research Council (CNR), Italy
Tabea Tietz, FIZ Karlsruhe and Karlsruhe Insitute of Technology, Germany
Eleonora Marzi, University of Bologna, Italy
Andrea Poltronieri, University of Bologna, Italy

**Students**
Aleksandra Simic, Norwegian University of Science and Technology, Norway
Alexane Jouglar, Université de Namur, Belgium
Anna Sofia Lippolis, University of Bologna, Italy
Anouk Michelle Oudshoorn, TU Wien, Austria

Antonis Klironomos, Robert Bosch GmbH and University of Mannheim, Germany

Arianna Graciotti, University of Bologna, Italy

Arnaldo Tomasino, Politecnico di Bari, Italy

Aya Sahbi, Université de Caen Normandie, France

Beatriz Olarte Martinez, Politecnico di Milano, Italy

Bohui Zhang, King's College London, United Kingdom

Caitlin Woods, The University of Western Australia, Australia

Celian Ringwald, Inria, France

Chiara Di Bonaventura, King's College London, United Kingdom

Cristian Santini, FIZ Karlsruhe, Germany

Daniela Milon Flores, Université Grenoble Alpes, France

Daniil Dobriy, Vienna University of Economics and Business, Austria

Dennis Sommer, University Innsbruck and Semantic Technology Institute, Austria

Disha Purohit, Technische Informationsbibliothek and Leibniz University Hannover, Germany

Ebrahim Norouzi, FIZ Karlsruhe, Germany

Eleonora Mancini, University of Bologna, Italy

Emetis Niazmand, Technische Informationsbibliothek and Leibniz University Hannover, Germany

Ensiyeh Raoufi, University of Montpellier, France

Francisco Bolanos, Open University, United Kingdom

George Hannah, University of Liverpool, United Kingdom

Giada D'Ippolito, Università degli Studi di Genova, Italy

Ginwa Fakih, Nantes University, France

Golsa Heidari, Leibniz University of Hannover, Germany

Hassan Hussein, Technische Informationsbibliothek Hannover, Germany

Heng Zheng, University of Illinois Urbana Champaign, United States

Inês Koch, Universidade do Porto and INESC TEC, Portugal

Ioannis Dasoulas, KU Leuven, Belgium

Joao Vissoci, Duke University, United States

Johanna Rockstroh, Universit of Bremen, Germany

Kerstin Neubert, German Federal Institute for Risk Assessment, Germany

Laura Menotti, Università degli Studi di Padova, Italy

Liam Tirpitz, RWTH Aachen University, Germany

Majlinda Llugiqi, Vienna University of Economics and Business, Austria

Manoé Kieffer, Nantes Université, France

Marco Di Panfilo, Free University of Bozen-Bolzano, Italy

Martin Critelli, University of Venice, Italy

Martin Böckling, University of Mannheim, Germany

Mary Ann Tan, FIZ Karlsruhe, Germany

Mikael Lindekrans, Linköping University and Ragn-Sells AB, Sweden

Mohammad Javad Saeedizade, Linkoping University, Sweden

Nicolas Lazzari, University of Bologna, Italy

Nicolas Ferranti, Vienna University of Economics and Business, Austria

Philipp Hanisch, TU Dresden, Germany
Raia Abu Ahmad, Deutsches Forschungszentrum für Künstliche Intelligenz GmbH (DFKI), Germany
Reham Alharbi, University of Liverpool, United Kingdom
Rita Sousa, LASIGE, Portugal
Roberto Barile, University of Bari, Italy
Ruben Eschauzier, Ghent University, Belgium
Sara Bonfitto, Università degli studi di Milano, Italy
Sefika Efeoglu, FU Berlin and TU Berlin, Germany
Soulakshmee Nagowah, University of Mauritius, Mauritius
Vidyashree Tarikere, Ghent University, Belgium
Vincenzo De Leo, Università degli Studi di Cagliari, Italy
Weronika Lajewska, University of Stavanger, Norway
Xinyue Zhang, King's College London, United Kingdom
Xuemin Duan, KU Leuven, Belgium
Yashrajsinh Chudasama, Technische Informationsbibliothek and Leibniz University Hannover, Germany
Zafar Saeed, Linkoping University, Sweden

# Abstract

The International Semantic Web Research School (ISWS) is a week-long intensive program designed to immerse participants in the field. It offers a blend of lectures and keynote presentations delivered by leading researchers, alongside a "learning by doing" teamwork program. Under the guidance of renowned scientists, teams tackle open research problems, culminating in the co-authorship of a white paper. ISWS' program is specifically designed to cultivate the next generation of semantic web researchers, emphasizing not only creativity and methodological rigor but also a sense of community and enjoyment.

This document reports a collaborative effort performed by ten teams of students, each guided by a senior researcher as their mentor, attending ISWS 2023. Each team provided a different perspective to the topic of creative AI, substantiated by a set of research questions as the main subject of their investigation.

The 2023 edition of ISWS focuses on the intersection of Semantic Web technologies and Creative AI. The recent emergence of generative AI has significantly impacted the field of AI. The public availability of Large Language Models (LLMs) marks a distinct break in technological history.

LLMs have demonstrated their potential as tools for creative tasks such as text, image, and music generation. This strength likely stems from their ability to learn from vast amounts of data and develop the capacity to abstract over it. However, several challenges remain. These include issues like hallucination (generating factually incorrect content), information retrieval, and the knowledge acquisition bottleneck (the difficulty of efficiently incorporating new knowledge). Additionally, concerns regarding transparency, traceability, copyright, trust, and ethics require further investigation.

Symbolic knowledge representations of the Semantic Web, in particular ontologies and knowledge graphs are instead very powerful at handling transparency, specialized knowledge, reasoning, generalization or abstraction. Combining these two approaches is not yet studied sufficiently, though.

ISWS 2023 explored various intersections between Semantic Web technologies and creative AI. A key area of focus was the potential of LLMs as support tools for knowledge engineering. Participants also delved into the multifaceted applications of LLMs, including legal aspects of creative content production, humans in the loop, decentralised approaches to multimodal generative AI models, nanopublications and AI for personal scientific knowledge graphs, commonsense knowledge in automatic story and narrative completion, generative AI for art critique, prompt engineering, automatic music composition, commonsense prototyping and conceptual blending, and elicitation of tacit knowledge.

As Large Language Models and semantic technologies continue to evolve, new exciting prospects are emerging: a future where the boundaries between creative expression and factual knowledge become increasingly permeable and porous, leading to a world of knowledge that is both informative and inspiring.

# Contents

## II  Human After All - AI Capabilities and People's Rights  38

## 4  A Decentralized Ecosystem to Empower Artists to have Ownership and Control over their AI-generated Art  39

## 5  ChatGPT, is this legal? Impact of the unintended misuse of AI solutions  48

## 6  KUBA: Knowledge Graph Generation through User-Based Approach  56

## III  The Sound of Silence - Commonsense and Implicit Knowledge in Large Language Models  65

# List of Figures

# List of Tables

# Part I

# A Kind of Magic - Generative Models and Art

# Chapter 1

# Harmonizing Creativity: Exploring the Role of Knowledge Graphs in Music Composition

MARTIN BÖCKLING, MARCO DI PANFILO, RUBEN ESCHAUZIER, GINWA FAKIH, EMETIS NIAZMAND, JOAO RICARDO VISSOCI, ANDREA POLTRONIERI, VALENTINA PRESUTTI

## 1.1 Introduction

Advancements in creative Artificial Intelligence (AI) have elicited significant transformations across several human creativity domains. Through the emerging generative AI approaches, systems are able to create new content such as image synthesis, music composition, and literature generation with minimal human interaction. For instance, in the realm of image synthesis, innovative models like DALL-E exemplify the ability to translate textual prompts into corresponding visuals with accuracy. Meanwhile, for text generation, transformer-based models such as GPT-3 or 4 have demonstrated unprecedented capabilities. These large language models (LLMs) leverage deep learning techniques to generate contextually relevant and coherent text, ranging from poetic verses to short stories and technical articles. Music composition has also witnessed progress with the release of models like AVIA, muBERT, Beatoven.ai , or JukeBox [56]. These models showcase the capacity to create musical compositions spanning various styles and genres, supplement incomplete symphonies, and accurately mimic or even originate voices (e.g. Murf.ai).

The rapid growth of generative artificial intelligence (GAI) has developed uncertainty and expectation that computers could substitute humans in complex tasks. However, ethical discussions on the use of GAI have been raised, including concerns about unintended plagiarism, dissemination of incorrect or poor-quality information, and even fabrication of false references. The progress in creative GAI extends beyond the mere automation of creative thinking. It heralds a new paradigm where AI operates as a creative collaborator, augmenting human creativity by offering fresh perspectives and possibilities driven by data and algorithms.

For instance, the use of creative GAI as a support tool to conduct textual writing, marketing (using chatGPT), and computational coding (using Github Copilot) has been associated with increments in time to complete the tasks when compared to control. However, while highly disruptive, the emergence of creative GAI has important limitations in relation to creativity-related tasks.

LLMs, creative GAI still fails to respond with high-quality outputs when fed prompts contain technical language and offer very minimal interactive and creative flexibility for domain experts. A possible solution is the infusion of knowledge through domain-specific Knowledge Graphs and Ontologies. Domain-specific ontologies have been used to improve GAI in image classification and show promise to address other creative areas, such as music composition. While this collaborative role of creative GAI is cross-cutting to all fields, we will focus on the field of musical composition as the area of our discussion.

To summarize, the main contributions of this work are as follows:

- We first show that current publicly available text-to-music approaches do not sufficiently use musical theoretic terms in prompts to generate music in a survey.

- Secondly, we show that we can manually translate natural language prompts to SPARQL queries to extract domain knowledge from the KGs and generate a dataset for further training.

- Furthermore we show, using a proof-of-concept model, that we can use LLM to translate natural language prompts into SPARQL queries on the KGs.

- Finally, we propose a process to directly infuse relevant knowledge in the KGs into the AI.

## 1.2   Research Questions

The empirical evaluation aims to formulate the following research questions:

A. Can existing creative generation models effectively use domain-specific technical language to guide the generation procedure?

B. How can we use symbolic knowledge bases to improve the creative generation model's understanding and use of domain-specific technical terms?

For the second question, we will divide our research approach into two working hypothesises.

A. Can we use LLMs to retrieve ground truth labels from the Harmony Memory (Harmory) [23] KG, given a prompt containing domain-specific technical language?

B. Can we use the knowledge contained in the Harmory KG to augment the generative music model's understanding of technical terms through knowledge embedding-based fine-tuning?

## 1.3 Related Work

Generative AI models have made remarkable contributions across diverse domains, revolutionizing creativity, innovation, and data generation. Their impact spans various fields including art and design, music, healthcare and medicine, gaming, and virtual reality. These models enable the creation of digital content such as images, music, and natural language. Notably, ChatGPT, DALL-E-2 [139], and Codex [46] have gained significant attention.

In the field of music, several studies have explored the application of machine learning techniques to generate, complete, and modify music pieces. One notable work in this domain is Jukebox, proposed by [56], which is a generative model capable of producing raw audio music with coherence extending to multiple minutes. Additionally, there have been successful attempts to generate high-fidelity music from text descriptions, as demonstrated by the work of [4]. This shows the advancement in generative models for music generations and hihglight the potential for AI-driven music composition. They learn distributions of musical sequences and use this to predict the most likely next element of a piece of music. Moreover, AIVA is another generative model that employs deep learning techniques to analyze and learn from music database to create music pieces in a variety of different styles and genres. While these endeavors have achieved success in generating music compositions, they may still yield unfavorable results when prompted with specialized requests.

On the other hand, numerous studies have put forward approaches for word embeddings, which involve representing words as real-valued vectors that capture their semantic meaning. In this representation, words that are closer together in the vector space are assumed to have similar meanings. Numerous works use word embedding based approaches on music to create embeddings for symbolic music data.

## 1.4 Resources

In our experiment, two Knowledge Graphs, Chord Corpus (ChoCo) and Harmory, and one generative model, muBERT, are used.

ChoCo[1] [24] is a Knowledge Graph built based on Polifonia Ontology Network (PON) [33] offers over 20,000 timed chord annotations extracted from scores and tracks. These annotations are enriched with semantic information from multiple data sources. They cover a wide range of genres and styles, providing resources for chord analysis and exploration. The derived annotations contain provenance data, metadata, authors of the annotations, identifiers, and links.

From the collection of harmonic annotations in ChoCo, a subset is used to build a KG of harmonic patterns called Harmory [23]. Leveraging harmonies within the Harmory KG can provide valuable support to the creative process of music composition. Harmory consists of harmonic patterns extracted from a large and heterogeneous musical corpus. Utilizing a cognitive model of tonal harmony, it segments chord progressions into meaningful structures, allowing harmonic patterns to emerge through the comparison of their harmonic similarity. They showcased the potential of Harmory in enabling transparent, explainable, and accountable applications for human-machine creativity.

muBERT is a generative music streaming app. It is trained on a large corpus of music-related text, which includes lyrics, music reviews, and music descriptions. By using muBERT in our work, we can instantly generate an audio track of a specific length from a user prompt.

## 1.5  Proposed approach

In this section, we outline, for each research question, the approach, evaluation, and results obtained.

*RQ1. Approach.* We conduct a user-based survey on the accuracy of creative GAI-generated music clip to the human-natural language prompt.

We create six musical clips (three-derived from the naive prompts and three from specialists). We ask six musicians to rate each musical clip with the question: "Please rate to what extent the clip reflects the requested prompt". Ratings are expressed using a likert type scale of four points, ranging from "Not at all" to "Perfect Match". We analyze the survey responses using descriptive statistics and divide between naive and expert-based prompts.

*RQ2. Approach.* We evaluate the feasibility of using domain-specific KGs to improve (music) AI in its ability to collaborate with human professionals (e.g. musicians). We test this methodology using two approaches:

*RQ2. Working hypothesis 1 - Approach.* For our first approach for the KGs integration with to GAI, we manually translate natural language prompts from musicians into SPARQL Queries to extract knowledge from the ChoCo KG. For this study, we report the translated query, the knowledge extracted and showcase the data available to be used to train or to validate the responses of GAI tools by improving their ability to respond to specialized prompts.

*RQ2. Working hypothesis 2 - Approach.* Our second approach for the KGs integration with to GAI is a combination of the musician natural language

---

[1]https://github.com/smashub/choco

| Prompt ID | Technical Prompt | Naive Prompt |
|-----------|------------------|--------------|
| P1 | Generate a series of chord progressions in the key of D major that utilize secondary dominants and modal mixture. | Generate a song similar to "Thriller" by Michael Jackson. |
| P2 | Provide me with a chord progression that primarily employs quartal voicings in the key of E major. | Generate a song similar to "Despacito" by Luis Fonsi. |
| P3 | Compose a piece in F# major, having the $I - V - vi - iii - IV - I - IV - V$ progression as a bridge. | Generate a song similar to "Seven Nation Army" by White Stripes. |

Table 1.1: The technical and naive prompts used in the exploratory survey of current generative music models.

prompt and the KG, through an automated SPARQL query generation.

A natural language prompt contains implicit information that can be addressed by a partial section of the KG by a query to the relevant section. Different from the first attempt, to avoid doing this query manually, we use a Seq2Seq LLM model to automate the query, which showed promising results in previous study [183]. However, it is important to state that there are three possible error situations that could occur in this approach. The first problem that is possible to occur is the wrong parameter selection from the prompt. In such a problem, the SPARQL query could fail due to a non-existent variable. The second problem that might occur is the wrong extraction of the value for a parameter. A user could in such case put information to the GAN that they did not want to have. The third possibility covers the general mistake of the Seq2Seq language model in using a proper SPARQL syntax and structure. In such case, the query of the SPARQL query would always fail.

An alternative approach to the extraction of the information through SPARQL queries, which is not addressed in this study but it is hypothesized based on our experiments, is the use of already extracted random walks together with the user prompt. By treating extracted triples from random walks as sentences, both the prompt and random walks over the KG can be embedded in the same space. Thus, we can use random walks that are semantically similar to the prompt as input to a generative model.

The overall approach can be found in Figure 1.1. In general, the proposed architecture orients itself on two papers related to GANs presented in section 7.4. During the training process, we propose to use a generative model like a GAN architecture, which is differentiated between a generator module and a discriminator. Instead of the random noise, we introduce the embedding derived from the Graph Neural Network, which learns incrementally a predefined subset of the KG. This conditional input for the GAN allows the GAN to include the semantics together with the relatedness of the KG structure in the generation. In past research, this approach has been shown to be beneficial for image generation [157]. Our proof of concept tests the Seq2seq approach and illustrates the steps to integrate with the GAI.

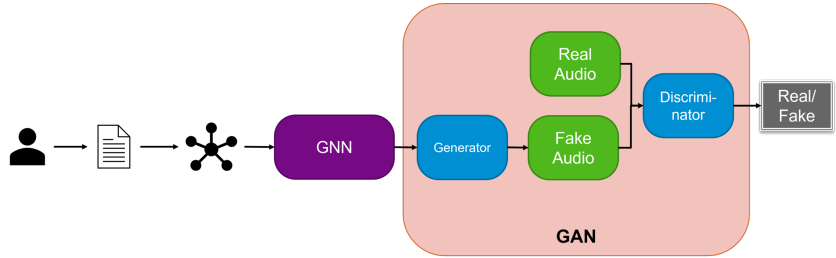

Figure 1.1: Proposed approach for GAN architecture with KG involvement

## 1.6   Evaluation and Results: Proof of Concept

**Answer to RQ1.** Qualitative user-centered feedback.To show that current creative GAI models do not effectively use domain-specific technical language, we conduct a mixed-methods survey with musical experts. We survey six participants, with one participant not completing the entire survey. On average, participants have 15 years of experience with musical theory. All participants play a form of instrument, with two participants having a classical musical theory background. In Table Table 1.2 we report the frequency of the four different possible answers in the survey. Furthermore, we report the average frequency of the answers for both technical and naive prompts. From the table, we find that current music generation techniques are not consistently capable of generating music that satisfies the prompt. Additionally, the prompts with technical terms perform worse on average when compared to the naive prompts. This indicates that the model can not use musical theory vocabulary to guide the creation of relevant music. This supports the need for methods that can improve the understanding technical musical terms in prompts. Thus, we will investigate the use of high-quality knowledge graphs to augment GAI.

| Response Option | Naive | | | | Technical | | | |
|---|---|---|---|---|---|---|---|---|
| | P1 | P2 | P3 | Average | P1 | P2 | P3 | Average |
| Not at all | 3 | 6 | 3 | 4 | 6 | 3 | 6 | 5 |
| Somewhat | 2 | 0 | 2 | 1.33 | 0 | 3 | 0 | 1 |
| Almost | 1 | 0 | 0 | 0.33 | 0 | 0 | 0 | 0 |
| Perfect | 0 | 0 | 0 | 0 | 0 | 0 | 0 | 0 |

Table 1.2: The results of our survey, with the frequency of four different possible answers possible.

```
PREFIX har:  <http://w3id.org/polifonia/harmory/>
PREFIX mf:   <http://w3id.org/polifonia/musical-features/>
PREFIX core: <http://w3id.org/polifonia/core/>

SELECT DISTINCT ?segment ?title
WHERE {?track har:containsSegment ?segment .
?track  core:hasTitle  ?title .
?segment  har:containsChordAnnotation  ? chord_sequence  .
?chord_sequence  mf:hasChord  ?chord ;
 mf:hasKey  ?key .
   FILTER (?key = "G#:min" ) .
   FILTER (CONTAINS( ?chord ,  " :maj7 ")  ||
   CONTAINS( ?chord ,  " :maj7 (#11)")  ||
   CONTAINS( ? chord ,  " :min7 ( b5 ) " ) )
} LIMIT 10
```

Figure 1.2: SPARQL Query to retrieve four segments to create a chord progression from track 'Ironic'.

**Answer to RQ2.** Symbolic knowledge infusion to creative generation in music GAI and Harmory KG to answer a set of prompts, the prompt "Create a chord progression in the key of G# minor that utilizes chords such as major 7th, dominant 7th sharp 11, and minor 7th with a flat 5." has been written to SPARQL query to interrogate the knowledge graph (Figure 1.2).

By traversing the Harmory KG, it facilitates granular information of the harmonic structure of songs. This traversal enables exploration of the harmonic segments of each song, identification of associated patterns for each segment, and the similarities between patterns and segments across different pieces. The results from the KG can be a valuable resource for training any LLMs. Harmory KG along with its ontology, provide structured and organized information, which can help improve the quality and relevance of the data used to train language models. It enables more accurate and context-aware responses, and assist in generating coherent and knowledgeable outputs.

To automatize the generation of SPARQL queries from user input, we performed a small experiment by fine-tuning a Seq2seq LLM. Due to time constraints, we used a single SPARQL skeleton and parametrized the input by replacing part of the query with values from a song dataset [147]. To test the quality of the output, we used the same skeleton as input and found that the LLM can reproduce a similar SPARQL query with minor errors. Due to limited computational resources and the previously mentioned time constraints, we are unable to translate a more general prompt to a SPARQL query. While the SELECT statement was properly generated, the WHERE statement was not. Nevertheless, this proof-of-concept shows that fine-tuning an LLM to generate valid SPARQL queries given more computational resources and a larger and more diverse dataset.

## 1.7 Discussions and Conclusions

In this paper, we show that current tools for generative AI to interact with musicians to create new songs are not yet sufficient to capture the semantic of technical prompt. Our survey shows that current generative AI tools are better in interpreting more general prompts than technical terms. The generated audio files to the general prompts were evaluated more correct than those from technical prompts.

The outlined approach in section 1.5, we introduced an approach implemented in image generation. During our research, the working hypothesis for future research is, that an adoption of the underlying generative architecture to the sound could be beneficial. For the entire architecture, it is important to state that for image generation the convolutional elements are based on squares. This aspect needs to be adapted for future work to fit the vectorized structure of audio files. Furthermore, the current approach involves GANs as the main generative model. Research has indicated that diffusion models outperform current GAN architectures [52].

For the retrieval of the KG for the generative model, it is shown in Figure 1.1 and section 1.6 that the generation of SPARQL queries can when executed on the KG lead to error situations. Therefore, we propose to use the second outlined approach to feed the structural information from the KG into the graph neural network (GNN). Nevertheless, it might be still beneficial for researchers to make the endpoint more accessible and train a specific dataset for the text to SPARQL translation using LLM. A similar diverse dataset as used in the paper from might introduce value for the querying of KGs.

# Chapter 2

# Can Machines Identify Great Works of Art?

ALEXANE JOUGLAR, ANTONIS KLIRONOMOS, ANNA SOFIA LIPPOLIS, DANIELA FERNANDA MILÓN FLORES, EBRAHIM NOROUZI, HENG ZHENG, AIDAN HOGAN

## 2.1  Research Questions

The proposed approach facilitates the ability of machines to identify the greatness of artworks. The assessment of this objective consists in the evaluation of the following research question: *Can machine learning techniques be used to identify great works of art based on different features?*

## 2.2  Semantic Web and Creative AI

In this paper, we consider how the Semantic Web can be applied for creative AI as whether machines can use data enriched by knowledge graphs (KGs) to improve their performance to identify the greatness of artworks, a subjective assessment norm often used by humans, through the objective properties pertained by the works.

## 2.3  Introduction

The aim of this work is to investigate the autonomous identification ability of great paintings by machines.

Despite numerous attempts in art history to address objective traits of the common understanding of "great" art by separating aesthetics (i.e. artistic value) and beauty (i.e. individual liking), there is no universally agreed-upon the definition of art [53, 54, 57]. A study conducted over fifty years ago shows that both humans and machines found it challenging to distinguish a Mondrian artwork and a computer-generated image [124]. The majority of the participants preferred the image generated by the computer. The intersection of technology and art can shape the cognitive networks of humans and impact society's attempt to fabricate new ideas and images [163].

The recent development of AI technologies improves the performance of autonomous art analysis [40]. Yet we still lack of assessment of the artworks by machines using humans' subjective criteria, such as the greatness of artworks.

Given the absence of a universal agreement upon the definition of "great" art, the approach presented in this paper considers the greatness of artworks based on Wikidata properties[1] related to artworks. Following [53, 57, 54], the concept of greatness in art can at least be characterized by four aspects:

A. **Normative feature**: Artworks can possess significant moral, political, and aesthetic power, contributing to the collective construction of meaning;

B. **Aesthetic interest feature**: level of appeal, fascination, or value associated with the aesthetic qualities or characteristics of an artwork. It is closely linked to popularity and the ability to captivate viewers, such as number of museum visitors;

C. **Rhetorical feature**: Often employed metaphorically, this aspect prompts audience engagement by leaving certain elements open to interpretation, encouraging viewers to actively participate in completing the missing pieces;

D. **Novelty feature**: Great artworks often exhibit elements of innovation, introducing fresh ideas, techniques, or perspectives that challenge conventional norms and offer new insights.

In this pilot analysis, we focus on the Wikidata property P6379, "has works in the collection". This property describes whether an artist has his works in the collection or not.

To narrow down our research scope, we deliberately choose the collections within the most visited art museums based on the Wikipedia webpage about the most visited art museums[2].

We use this property to identify the greatness of the artworks by assuming that if the creators of them have works collected by top visited museums, then it means that these artists are recognizable by people, and thereby in a sense that their works are considered to be great.

This exploration enables us to assess the significance of artworks based on their association with specific collections, shedding light on their cultural and historical value within the broader art landscape.

The research conducted thus far will contribute as follows:

---

[1] https://www.wikidata.org/wiki/Wikidata:Main_Page.
[2] https://en.wikipedia.org/wiki/List_of_most-visited_art_museums

A. Leverage data enriched by the domain-specific knowledge graphs, focusing on the properties inherent to the artworks.

B. Using KGs and a single objective criterion (for example, whether the artist has artworks in a top 10 visited art museum).

**Outline**

We organize the paper as follows: Section 2.4 presents the state of the art. Section 2.5 discusses the resources. Sections 2.6 and 2.7 show the our method and evaluation of it. Section 2.8 discusses and concludes the paper. All the resources are available on Github[3] and licensed under GNU[4].

## 2.4  Related Work

Traditional approaches to automatic art analysis use hand-crafted features and Machine Learning algorithms [13, 32, 93]. However, these approaches face challenges in acquiring explicit knowledge of the attributes associated with specific artists or artworks. Such knowledge is often based on implicit and subjective experiences that even human experts find difficult to articulate [39, 145].

In contrast to these approaches, representation learning has been proven to be effective in various Computer Vision tasks, such as applying deep neural networks to art analysis [91]. Numerous works have explored the use of Deep Learning (DL) techniques, including both single-input [45] and multi-input models [158], for predicting artwork attributes based on visual features.

Additionally, researchers in this field have investigated other areas of interest such as visual link retrieval [37, 151], object detection [151, 73, 76], and near-duplicate detection [156].

Predicting artwork attributes using visual information is a challenging task. However, this can be overcome with the inclusion of contextual information alongside visual features. For instance, utilizing KGs to represent contextual information and integrating it into DL models. Garcia et al. [69] introduced the ContextNet framework, which combines a multi-output Convolutional Neural Network (CNN) for attribute prediction based on visual features with an encoder that leverages non-visual information from artistic metadata encoded using a KG. However, this approach has limitations due to the availability of metadata and a lack of consideration for other distinguishing features of available artworks. These limitations are also reflected in the design of the ontology.

To address these limitations, Castellano et al. [36] proposed to use external knowledge from Wikipedia, which offers a vast amount of structured information. By incorporating this external knowledge source, it is possible to enhance the understanding of artwork attributes by overcoming the lack of domain information for new artworks and considering relationships between artists, such as

---

artistic influence. However, in the resulting graph, entity linking is incorrectly mapped to artists instead of artworks, and several other mappings are missing. To address this issue effectively, we need to expand both the dataset and the ontology. By incorporating distinguishing features specific to artworks, we can enhance the accuracy and comprehensiveness of the mappings.

Analyzing artworks from a subjective perspective still remains an open topic, as subjective evaluation often involves complex human interpretation. This problem has been addressed in [16, 39].

## 2.5 Resources

**ArtGraph.** The main dataset used in this work is ArtGraph, a KG in the art domain, based on WikiArt [133] and DBpedia [170]. The dataset represents and describe concepts related to artworks [35]. The graph is made up of 135,038 resources out of which 116,475 are artworks from 186 periods. The dataset contains information about the date, title, genre, image, dimensions, URLs to Wikipedia, Wikidata, and DBPedia, and name and birth date of the author. The ArtGraph KG is accompanied by an ontology built around two main classes: *Artist* and *Artwork* (Fig. 2.1).

For the purpose of this initial paper, the utilized components of ArtGraph are:

- *createdBy* object property: connects artworks with artists

- *wikipedia_url* datatype property: provides the Wikipedia webpage URL of an artist [5]

- *image_url* datatype property: has the URL of painting images

**Wikidata.** Wikidata is a centralized knowledge base (KB) that stores structured data for Wikipedia and other Wikimedia projects, providing a common data source for information shared across multiple language editions. It supports the querying and retrieval of specific data through its triple-based structure.

This paper focuses on the below components of this KB:

- *has works in the collection* (*P6379* ) object property: connects an artist with the collection that has works made by him

- *schema:about* object property: inherited by schema.org, this property links an artist's Wikipedia webpage URL (i.e. https://wikipedia.org/wiki/Pablo_Picasso) with the corresponding Wikidata URL (i.e. https://www.wikidata.org/wiki/Q5593)

- *art museum* (*Q207694* ) class: from this class, the top 11 most visited art museums [106] were chosen: *Louvre Museum* (*Q19675* ), *Vatican Museums* (*Q182955* ), *British Museum* (*Q6373* ), *Tate Modern* (*Q193375* ), *National*

---

*Museum of Modern and Contemporary Art, South Korea (Q484054), Musée d'Orsay (Q23402), National Gallery of Art (Q214867), Metropolitan Museum of Art (Q160236), National museum of modern art (Q1895953), Hermitage Museum (Q132783), National Gallery (Q180788).*

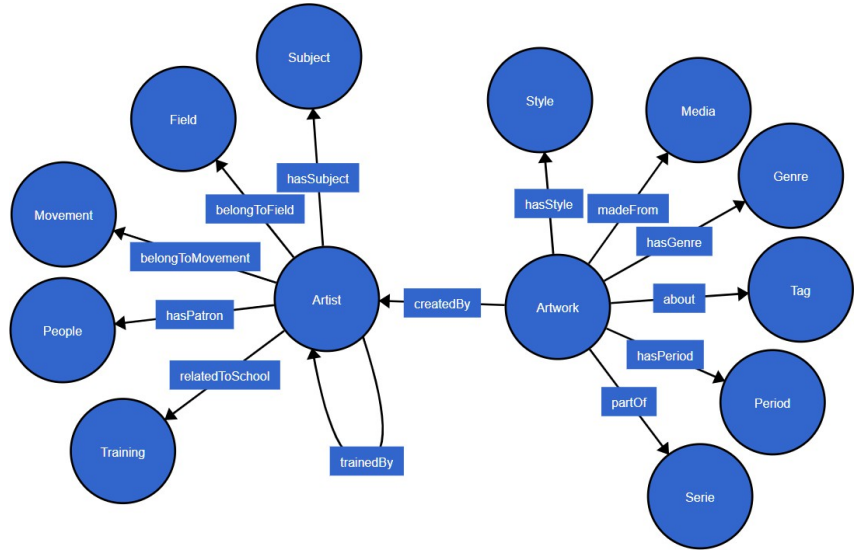

Figure 2.1: ArtGraph ontology.

## 2.6 Proposed approach

### 2.6.1 Data processing

The primary data source is the ArtGraph KG from which the *Artwork* instances were filtered using SPARQL to keep only those that were connected to an *Artist* instance via the *createdBy* property. Afterward, linking was performed between ArtGraph and Wikidata using the *wikipedia_url* data property of ArtGraph to get the equivalent Wikidata URL from the tail entity of triples with the format: *wikipedia_url schema:about wikidata_url*. Eventually, the painting images were downloaded using the ArtGraph *image_url* data property.

### 2.6.2 Criteria for automatic labeling

The binary labeling of paintings as 'great works of art' or not was performed in an automated fashion by taking into account the cardinality of the *has works in the collection* Wikidata object property. In particular, for every painting that was included in the dataset: if its artist has published at least one artwork in the collection of at least one of the ten most visited museums, then the painting is considered a 'great work of art'. Otherwise, not. By utilizing this characteristic, we gauge the magnificence of artworks by postulating that if renowned museums have acquired pieces from their creators, it implies that these artists are widely acknowledged, thereby implying the exceptional quality of their works.

### 2.6.3 Classification model

The proposed approach is shown in Fig. 2.

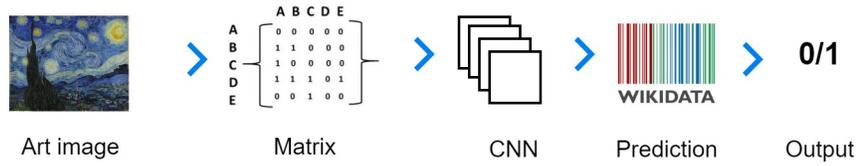

Art image      Matrix      CNN      Prediction      Output

Figure 2.2: Proposed workflow.

## 2.7 Evaluation and Results: Use case/Proof of concept - Experiments

The CNN model is used to train on the dataset of 560 paintings from WikiArt and 240 paintings are used for the test. The whole dataset contains 685 paintings classified as great works of art and 115 paintings as not great works of art. We used two dense layers (128, 1), and the Sigmoid activation function is used with the Adam optimizer. Binary cross entropy is used as a loss function.

Figure 2.3 shows the accuracy of the correct classifications that a trained machine learning model achieves.

## 2.8 Discussion and Conclusions

The presented approach allows machines to identify the greatness of artworks, which has not been well-discussed in the literature, using techniques that combine knowledge-based and learning-based AI approaches. The greatness feature of artworks is based on subjective human assessment, which is considered to be hard for autonomous identification. We, therefore, interpret this subjective assessment feature by objective properties from Wikidata, such as whether the artists behind the works have works collected by top visited museums.

### 2.8.1 Limitation

**The property we use attributes to artists** The property we used in this paper, i.e., *has artworks in the collection*, is a property attributed to the artists instead of the artworks themselves. Hence our results may identify the style of the artists are great instead of a specific artwork. In the future, we are going to find and enrich the dataset with the mappings of the artworks from the existing knowledge graphs.

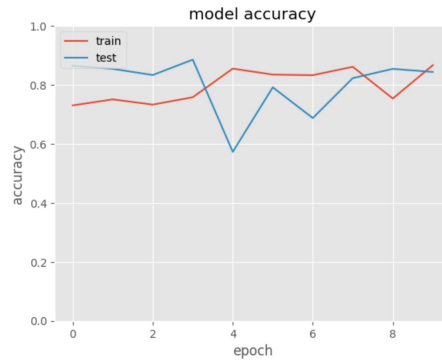

Figure 2.3: Model accuracy.

**Data enrichment** According to our analysis, we found that the ArtGraph framework provides inadequate input due to its limited scope: primarily focusing on mapping the artists behind the artworks from Wikipedia instead of the artworks themselves. Similarly, Wikiart also fails to provide comprehensive resources. Therefore, we recommend enhancing the existing WikiArt KG by incorporating a broader range of data, including not only the artists but also the artworks themselves and other relevant Wikidata features.

To achieve this, we propose developing a novel KG that encompasses the entirety of WikiArt's data. This new graph should facilitate the mapping of artworks, their respective attributes, and the artists involved. This may need a customized ontology tailored to the specific requirements of the project. The proposed knowledge graph is necessary for achieving a comprehensive and robust artistic evaluation system.

**Characterization of the greatness** As the concept of "greatness" encompasses various attributes, we need to consider more Wikidata properties that align with the definitions outlined in [53, 54, 57] in order to achieve better results. For a more comprehensive evaluation of artistic greatness, we can utilize the properties described in Table 2.1. These properties encompass various aspects of an artwork, including its artistic value, institutionalization, political significance, conveyed message, and presence both online and offline. By considering these properties, we can conduct a more thorough assessment of artistic greatness.

Table 2.1: Other features we considered can be used to identify great artworks.

| Feature | Property code | Property name |
|---|---|---|
| Normative | P276 | Location |
|  | P195 | Collection |
| Normative | P135 | Movement |
|  | P793 | Significant event |
| Aesthetic interest | P5008 | On focus list of Wikimedia project |
|  | P1343 | Described by source |
|  | P973 | Described at URL |
|  | P1433 | Published in |
|  | P2047 | Duration |
| Rhetorical ellipsis | P1257 | Depicts Iconclass notation |
|  | P921 | Main subject |
| Novelty | P166 | Award received |

## 2.8.2   Future work

We are going to enrich the current dataset with the DBpedia and Wikidata knowledge graphs. Furthermore, the hyperparameters of the machine learning model should be optimized. Additionally, we are going to investigate which features of works of art influence most of the models' predictions and identify which forms of art or greatness are difficult to recognize.

## 2.8.3   Conclusions

In this paper, we propose a machine-learning method using enriched data to identify the greatness of artworks. We trained the machine for understanding subjective human evaluation criteria by using objective properties in Wikidata. The limitations we showed point to the future direction of our work.

# Chapter 3

# Are Large Language Models the End of Knowledge Graphs? An Experiment in Story Completion

ARIANNA GRACIOTTI, CHIARA DI BONAVENTURA, SARA BONFITTO, DENNIS SOMMER, MANOÉ KIEFFER, MAJLINDA LLUGIQI, JAN-CHRISTOPH KALO, FRANK VAN HARMELEN

## 3.1 Introduction

A revolutionary class of new AI systems, known as Large Language Models (LLMs) [27], displays emergent behaviour and provides impressive abilities to generate text, images, computer programs and more, in a seemingly creative manner. Since these LLMs are very large transformer models [169], all information they contain is in latent form, distributed across hundreds of billions of nodes in the neural network. The impressive performance of LLMs across a wide range of domains and tasks [18] raises the question:

*Is there still a role for explicit, symbolically represented knowledge in the era of LLMs?*

This question is particularly urgent for the Semantic Web community, which has invested substantially in the development of large volumes of symbolically represented knowledge in the form of Knowledge Graphs (KGs). There has been speculation that KG's can enhance the performance of LLMs for reducing hallucination, providing explanations, and zero-shot learning [1].

---

[1] https://chat.openai.com/share/20986967-29fa-4578-8caf-3d5e61b9edb7

Unlike these speculations, we set out to answer the above question in the context of *Creative AI*. The Creative AI setting would seem to favour the probabilistic prediction processes of the LLMs over the rigid logical reasoning of KGs. This makes it a good experimental setting for the above question, since it stacks the odds against KGs, so being able to prove the added value of KGs in the context of Creative AI would provide strong evidence indeed. Within the domain of Creative AI, we choose the task of story completion. Completing a short story clearly falls in the domain of *Creative AI*, it is a task at which LLMs can be expected to perform well (so making it not obvious whether KGs can contribute), and there exist validated corpora for this task.

## 3.2   Research Questions

We operationalise our overall question on the role of KGs in the era of LLMs in the following specific research questions:

*RQ1: Does incorporating relevant knowledge from KGs enhance the ability of LMs to generate a plausible story completion?*
We expect that the performance of LLMs on the story completion task will vary with the size of the LLM (the number of trainable parameters in the model). This opens up the question if the degree to which KGs can support LLMs in the story completion task increases for smaller LMs. This leads to the second research question:

*RQ2a: Does the degree to which KGs improve the performance of LMs decrease with increasing size of the LM?*

*RQ2b: Do less complex LMs enhanced with knowledge graphs deliver comparable performance to large models?*

## 3.3   Semantic Web for Creative AI

Within the domain of Creative AI we choose the task of story completion. Completing a short story clearly falls in the domain of "Creative AI". Furthermore, it is a task at which LLMs can be expected to perform well, so making the question of whether KGs can contribute to LLMs not simply a foregone conclusion. Finally, it is one of the few tasks in the domain of Creative AI for which there exist validated corpora that can be algorithmically validated. Our hypothesis is that

H1:  Large Language Models (LLMs) will perform so well on Creative AI tasks that injection of KG's will not further improve their performance

H2:  Small Language Models (SLMs), which we expect to perform significantly worse on this Creative AI task, will benefit from the injection of KG's for their performance on story completion.

In summary, we use the domain of Creative AI as a touchstone to determine the value of symbolic representations (Knowledge Graphs) over and above the distributed representations of language models.

## 3.4   Related Work

The field of story completion has witnessed remarkable advancements in recent years, driven primarily by the growth of LMs and their ability to generate coherent text [84]. Researchers have explored various approaches to further enhance the performance of LMs in story completion tasks. One promising direction is the incorporation of KGs, which offer structured representations of domain-specific information. By leveraging the rich connections between entities, concepts, and relationships encoded in KGs, researchers have aimed to enhance LMs' ability to generate more coherent and semantically accurate story completions. Several studies have demonstrated the efficacy of integrating KGs into LMs, resulting in improved narrative coherence, increased factual consistency, and enhanced contextual understanding.

Chen et al. [44] proposed a neural network model that integrates three types of information: narrative sequence, sentiment evolution, and commonsense knowledge from the ConceptNet. By combining these sources of information, they achieved improved performance compared to state-of-the-art approaches on the ROCStory Cloze Task dataset.

Guan et al.[75] extended the GPT-2 by incorporating external commonsense knowledge from ConceptNet and ATOMIC. By post-training the model on knowledge examples derived from these sources, they effectively enhance the long-range coherence of the generated stories. Additionally, to ensure the generation of plausible narratives, they introduce a classification task that distinguishes true stories from automatically constructed fake stories. This auxiliary task encourages the model to capture causal and temporal dependencies between sentences, leading to improved inter-sentence coherence and reduced repetition. Through comprehensive automatic and manual evaluations, the experimental results demonstrate that their model outperforms strong baselines in terms of logicality and global coherence, thereby generating more reasonable stories.

Yang et al. [182] presented an alternative method of incorporating information from a KG into the prompt. They started by extracting keywords from the context story and linking them using only text-based information to construct a KG. Furthermore, they incorporated knowledge from external sources such as DBPedia, ConceptNet, and WordNet. The resulting KG was used to extract triples, which were then injected into the prompt. Notably, their approach differs from ours as it lacks information regarding the classes of named entities.

Going beyond all the above, our proposed approach focuses particularly on the complexity of the LM. We compare the effect of incorporating information from KGs in LMs of different sizes.

## 3.5 Resources

**FLAN-Alpaca.** We chose to utilize instruction-fine-tuned language models[2], trained on the Alpaca [160] and FLAN [50] instruction datasets as our SLMs. Alpaca is available in different sizes (base, large, XL, XXL) making it possible to test our hypothesis. We decided to use Flan-Alpaca-Base (220M parameters), Flan-Alpaca-Large (770M parameters), and Flan-Alpaca-XL (3B parameters) to represent three different SLMs.

**GPT-3.5.** In addition, we are using GPT-3.5-turbo[3] in order to compare the results of our KG-enhanced SLM to an LLM. The models will be tested on both ROCstories [116] and Possible Stories [15]. The model also known as text-davinci-003, is an advanced language model developed by OpenAI. It belongs to the GPT series and utilizes machine learning to generate human-like text. Trained on diverse internet text, it doesn't have access to specific training documents or personal data unless provided during conversation. Like its predecessors, GPT-3.5 employs transformer architecture, which revolutionizes sequential data processing by eliminating the need for sequential computation. With a size similar to GPT-3 (175 billion parameters), it excels at generating coherent and contextually relevant sentences. While expected to outperform previous models in language tasks, exact improvements are challenging to quantify without specific details or direct comparisons.

**ROCstories.** The ROCstories dataset contains context stories of four phrases and two possible candidate phrases [116]. One of the candidates is considered a good ending of the story, the other a wrong ending. The task of story completion then comes down to choosing which of the candidate phrases is the good ending.

**Possible Stories.** The Possible Stories dataset contains a context story of four phrases [15]. For each context story, the dataset contains four candidate answers and a set of two questions. The good answer for a given story then depends on which question is asked. The task of story completion is then to find the good answer out of four candidate answers given a context story and a question.

## 3.6 Proposed approach

Our approach to enhancing a story completion prompt with the NERC information and WSD knowledge of WordNet can be described in three steps: (i) The extraction of named entities and their corresponding classes from the context stories. (ii) The disambiguation of the rest of the context story. (iii) The injection of the acquired knowledge within the prompt.

---

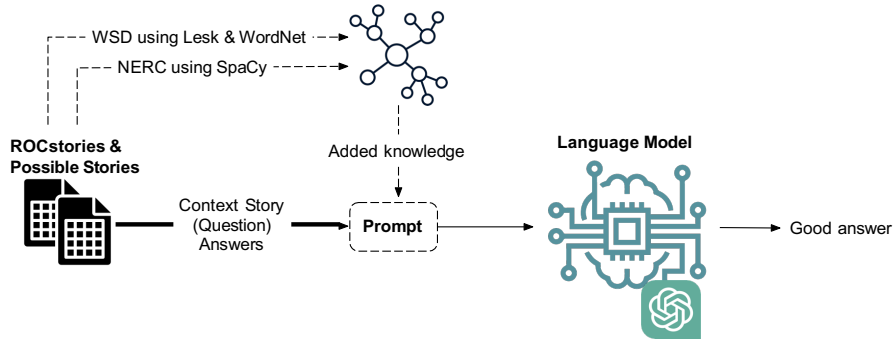

Figure 3.1: Approach to create KG-enhanced prompts

We compare this knowledge-enhanced approach to baseline experiments in which the prompt does not report any injected knowledge.

**Baseline experiments** In the baseline experiments setting, we create a plain prompt to condition the model to predict the correct story completion. An example of the prompt sent to the model in the baseline experimental setting is reported in 3.6.

```
A is the beginning of a story. Which sentence correctly completes
    the story, 1 or 2?
A= Rick grew up in a troubled household. He never found good support
    in family, and turned to gangs. It wasn't long before Rick got
    shot in a robbery. The incident caused him to turn a new leaf.
1= He is happy now.
2= He joined a gang.
```

**Knowledge Injection Experiments** For Named Entity Recognition and Classification (NERC), we extract the named entity from each context story using SpaCy[80]. Spacy is an open-source natural language processing library that provides efficient and powerful tools for text analysis. Spacy allows the recognition and classification of the named entity present in the story via a fast statistical approach. From it, we get a set of named entities and their corresponding classes. Examples of such information can be seen in lines 1 and 2 below.

For Word Sense Disambiguation (WSD) we use the Lesk algorithm[103]. We rely on WordNet as Knowledge Base, selecting word senses from it. In order to achieve WSD, the algorithm compares the candidate definitions of a specific word. The candidate definitions are compared to the definitions of the words in the corresponding surrounding context. The one with the highest overlap is chosen as the most appropriate definition. An example of the knowledge extracted can be seen in lines 3, 4, and 5 of the following example:

```
'Sarah has_label People, including fictional'
'years has_label Absolute or relative dates or periods'
'like has_definition prefer or wish to do something'
'different has_definition marked by dissimilarity'
```

```
'dreaming has_definition experience while sleeping'
```

In this step, we create the prompts based on the knowledge extracted in the two previous sections. For both ROCstories and Possible Stories, the prompts are composed of: the context story, the candidate answers, and the injected triples. For Possible Stories, we also add the question. An example of a prompt for ROCstories can be seen below. The prompts for Possible Stories are designed analogously.

```
A is the beginning of a story. Which sentence correctly completes
    the story, 1 or 2? You must answer 1 or 2, no other answers
    allowed.
A=Sam loved his old belt. He matched it with everything.
    Unfortunately he gained too much weight. It became too small.
1=Sam went on a diet.
2=Sam was happy.
When answering, please consider the following knowledge:
['Sam has_label People, including fictional', 'loved has_definition
    have sexual intercourse with', 'old has_definition just
    preceding something else in time or order', 'belt
    has_definition fasten with a belt', 'matched has_definition set
    into opposition or rivalry', 'Unfortunately has_definition by
    bad luck', 'gained has_definition reach a destination, either
    real or abstract', 'weight has_definition a unit used to
    measure weight', 'small has_definition have fine or very small
    constituent particles'].
```

**Few Shot Approaches** In addition to the zero-shot approaches described in the previous paragraphs, we tried three different few-shot learning approaches to provide auxiliary information to the model in the form of single-shot example. Precisely, we prompted the model as follows: *(i)* single-shot example and story augmented with their related triples, *(ii)* only single-shot example augmented with its related triples, and *(iii)* single-shot example and story without any triple injected.

## 3.7 Evaluation and Results

We compared our proposed approach to a baseline composed of the three Flan-Alpaca models (base, large, XL) as the representative of SLMs, and GPT-3.5-turbo as the representative of LLMs. The evaluations were done on two different story completion datasets: ROCstories and Possible Stories. For both, the relevant information was given as the prompt.

When injecting background knowledge we added the context information obtained from the context story and WordNet [111] using the Lesk algorithm and SpaCy in the form of triples. This knowledge follows the actual prompt. We tested our approach on 1571 ROCstories and on 3414 Possible Stories. The models were used as it is and not trained by us.

Table 3.1 compares the accuracy of our approach and a baseline method across four levels of model complexity in two datasets, in a zero-shot experimental setting. On ROCstories dataset, for Flan-Alpaca models, the baseline approach achieves accuracy scores of 0.78, 0.89, and 0.98 for the Base, L, and XL levels, respectively whereas the most complex model tested, namely GPT3.5, achieved 0.967 accuracy. Our proposed approach when using Flan-Alpaca-Base achieves an accuracy score of 0.67. On Possible Stories dataset, the baseline achieves an accuracy score of 0.42 for the Flan-Alpaca-L model.

Table 3.1: Comparing the accuracy of our approach to the baseline across four levels of model complexity in two datasets in zero-shot experimental setting.

| | | Flan-Alpaca | | | GPT |
|---|---|---|---|---|---|
| **Dataset** | **Model** | **Base** | **L** | **XL** | **3.5-turbo** |
| ROCStories | Baseline | 0.78 | 0.89 | 0.98 | 0.967 |
| | Our approach | 0.67 | | | |
| Possible Stories | Baseline | | 0.42 | | |
| | Our approach | | | | |

Table 3.2: Comparing the accuracy of our approach to the baseline in a few-shot experimental setting.

| | **FS (Baseline - no knowledge)** | **FS (partial knowledge)** | **FS (all knowledge)** |
|---|---|---|---|
| Flan-Alpaca-Base | 0.76 | 0.7 | 0.58 |

In Table 3.2, we present a comparison of the accuracy achieved by our approach and the baseline in a few-shot experimental setting according to Flan-Alpaca-Base model, focusing on different levels of knowledge injection. Specifically, we consider three conditions for Few-Shot learning (FS): FS (Baseline - no knowledge), FS (partial knowledge), and FS (all knowledge). We refer to partial knowledge when injecting knowledge into the single-shot example only, whereas we use the term all knowledge when the knowledge is injected both in the single-shot example and input story. The baseline achieves an accuracy score of 0.76 in the FS (Baseline - no knowledge) condition, followed by 0.7 in the FS (partial knowledge) condition and 0.58 in the FS (all knowledge) condition.

## 3.8 Discussion and Conclusions

**Discussion of Results.** The largest language models reach performance close to perfection in the story completion task on the ROCStories in a zero-shot setup and with no symbolic knowledge injected. The smallest showed room for improvement so we tried injecting knowledge into them.

We injected knowledge in the form of triples that were added to the prompts sent to the smallest LLMs. This knowledge encompassed Named Entity Recognition and Classification (NERC) and Word Sense Disambiguation (WSD) information extrapolated from the input text (a story from ROCStories). NERC was performed using SpaCy. For WSD, we relied on WordNet as a Knowledge Base.

The injection of knowledge always and consistently caused a significant *decrease* in performance on the task of story completion.

Our experiments show that the amount of knowledge injected matters: in the few-shot learning setting, the injection of all WSD and NERC triples significantly decreased accuracy when providing triples extracted from both the single-shot example and the input story. The performance drop was lower when we injected triples related only to the single-shot example but not to the input story for which the model had to predict a completion.

From the experiments, we learned that knowledge injection can introduce noise so we believe that in-depth investigation of the type of knowledge (in both zero-shot and few-shot learning settings) and its degree of injection (in few-shot learning settings) is crucial to use KGs as an auxiliary tool for language models.

**Conclusion.** Our first hypothesis was easily confirmed in our experiments: LLMs have a near-perfect performance on the story completion task. However, our experiments clearly falsified our second hypothesis: the injection of knowledge always and consistently caused a significant *decrease* in performance on the task of story completion. This conclusion raises a number of questions to be investigated in future work.

*Types of knowledge:* Our injected knowledge was very linguistic in nature (NERC and WSD). Perhaps the injection of more factual or commonsense knowledge would be more effective in boosting the performance of small language models.

*Amounts of knowledge:* Our experiments showed that the amount of injected knowledge also plays a crucial role. Again, experiments on the role of this parameter are left for future work.

*Nature of the task:* Our experiments are limited to choosing among two predefined completions. Future work should investigate the more creative task of actually *generating* a story completion. This would locate this work even more deeply in the domain of Creative AI.

# Part II

# Human After All - AI Capabilities and People's Rights

# Chapter 4

# A Decentralized Ecosystem to Empower Artists to have Ownership and Control over their AI-generated Art

ROBERTO BARILE, YASHRAJSINH CHUDASAMA, NICOLAS FERRANTI, GOLSA HEIDARI, VIDVASHREE TARIKERE, ARNALDO TOMASINO, JOHN DOMINGUE

## 4.1   Research Questions

**RQ1:** How can we guarantee that artist collectives have full ownership and control of their work? As they create, advertise, and sell their work?

- **RQ1.1:** Can decentralized technologies, such as distributed ledgers or personal data pods support RQ1?

- **RQ1.2:** Can semantic technologies be used to represent generative AI concepts and enable semantic search over generative AI artistic output and their associated prompts?

## 4.2   Semantic Web for Creative AI

The technologies falling under the umbrella term *Generative AI*, e.g. ChatGPT and DALL-E 2 by OpenAI [127], showed great potential in performing creative tasks according to user-provided prompts, e.g., storytelling or painting, so artists around the world started to notice them and it quickly became common to involve them in the creation of art pieces. Artists questioned, even in legal actions [14], that AI tools are not generating "original" art pieces, but emulating human-created art and pasting it in a sort of collage. The artists who unwittingly helped train the tools want to be financially compensated for their work. Artists also use to form artist collectives, they are fundamentally collaborative platforms where artists converge to create, showcase, and often promote their work. They function as cooperative entities that enable shared resources, mutual support and collective decision-making, typically oriented around a common philosophy, political view and aesthetic sense. Question arises about collectives that employ generative AI in their work, for instance the A.I. collective [5] an interdisciplinary artist collective that stands at the intersection of art, science, society and technology. It is not well defined who the real owners of the product are, we claim that the artists are because is on their behalf to creatively and effectively guide the interaction with the tool with appropriate creative prompts, artists own prompts and we claim that owning the prompt implies owning the result.

Artist collectives want to be empowered and have real ownership and control of their work. Ownership also means that it must be possible to buy and sell in a marketplace application such artistic properties, like prompts, and the results of prompt feeding. Successful implementation of a marketplace heavily depends on the availability of an effective search engine.

Artists, in particular musical ones, complain about centralized platforms, for instance, there is a lot of criticism towards the Spotify platform, mainly because of low payouts [78], for this reason it is also crucial, to achieve real ownership, that platform for artists have no central authority in order for collectives to not depend on a centralized platform and rely on a self-governed ecosystem.

This scenario can benefit from semantic technologies and from decentralization. Semantic technologies can enhance the marketplace implementing semantic search based on a knowledge graph, while decentralized technologies can help in cutting out the central authority.

## 4.3   Introduction

Art has long been recognized as a powerful means of creative expression and a reflection of human culture. Artists invest their time, energy, and emotions into the creation of their works, giving birth to unique pieces that capture the essence of their vision. These works hold immense personal and societal value, often becoming an integral part of our cultural heritage.

However, with the rise of generative AI, the dynamics surrounding artistic creation and ownership have begun to undergo significant transformations. Generative AI systems have demonstrated remarkable capabilities in producing art that can mimic or even surpass human creativity [139]. While this technological advancement brings forth exciting possibilities, it also raises profound concerns among artists who fear losing control over their creations [17].

Many artist collectives around the world have expressed concern that generative AI technologies, such as DALL-E, MidJourney, and ChatGPT violate their rights. These technologies generate high-quality images based on a given prompt, frequently without the artists' consent or awareness. Infringing on artists' collective intellectual rights causes economic and psychological harm. Some real-world examples from the perspective of the artist collective are: (1) in the UK a renowned media company, Getty Images [128], filed a judicial proceeding arguing that Diffusion AI technology is used as training data millions of images from the Getty platform without any consent. (2) Jon Lam [88], a Canadian advocate, has been influential in protecting artist rights against AI security breaches. He highlights the problem on social media, trying to engage artist collectives, researchers, and creators on the impact of generative AI. (3) As reported by BBC News [89], Open AI has accomplished a notable feat by successfully reproducing the viral song of the Beatles band, utilizing the voice of a singer as their basis.

The rights of artists are fundamental to the preservation of creative integrity and the sustainability of artistic communities. Historically, copyright laws and intellectual property frameworks have provided artists with legal protection, granting them exclusive rights over their works. These rights include the capacity to reproduce, distribute, and control derivative uses of their works. However, the rapid growth of generative AI has called into question traditional concepts of authorship and ownership, leaving artists dissatisfied with the status quo.

In this paper, we aim to address some of the presented issues faced by artists. Our main contributions are (1) a decentralized architecture to empower artists in controlling AI-generated works of art, comprising personal data pods, and ownership transactions registration on Blockchain; (2) an RDF model that reuses existing art terminology and comprises new terms to describe ownership and control of AI-generated art.

The remainder of the paper is as follows: Section 4.4 presents decentralized and semantic technologies for copyright protection of art. Section 4.5 and 4.6 introduce the architecture and modeling proposed to preserve ownership over AI-generated art. Lastly, Section 4.7 outlined our conclusions and future work.

## 4.4 Related Work

**Blockchain for copyright protection.** Blockchain [118] can be considered as a data structure where blocks of transactions get appended in an immutable chain fashion, thanks to the use of cryptographic techniques used to link blocks. Each ledger is distributed and synchronized among the participating nodes through peer-to-peer protocols based on consensus. Common approaches used for the management of intellectual properties in the context of blockchain are based on the use of smart contracts and external storage infrastructure, particularly IPFS. Design schemes [108, 109] that uses perceptual hash functions to generate hashes for the image as its digital signature and an infrastructure based on blockchain and IPFS. Images are watermarked and smart contract are used for detection of copyright infringements of new inserted pieces of art. These proposals are similar to ours, but we enhance the protection of data privacy and access control by using secure storage based on *Solid* pods and ODRL rules [126]; a vocabulary to describe the digital rights of the owners. Furthermore, Ethereum is being used also to manage authorization and copyright for watermarked music content [187], inserted into IPFS. The blockchain stores hashes of sound files. Ethereum is also utilized as a payment for gas prices, suggesting a decentralized marketplace similar to what has been proposed for implementation in this work.

**Storing data with Solid pods.** *Solid* [146] ecosystem fosters the separation of data from applications making the users the data owners with the ability to decide how it is used. *Solid* pods can store RDF and non-RDF data with access control policies. Every user storing data in *Solid* pods is identified by WebID [11] and is used for authentication with SolidOIDC [51] and Web Access Control (WAC) [174]. By utilizing Solid in the context of art ownership, artist collectives can retain full control over their creative works, ensuring that third parties cannot exploit or consume their art without appropriate access control.

Göndör et. al. [72] propose a blockchain technology, Blade, can be used as a decentralized and distributed ecosystem. Blade uses public-private key pair for authentication implemented with smart contracts using Ethereum. Blade has APIs for basic operations and it allows users to define access control for their own data. The data is stored in a database and it is only accessible by owners and other blade users with appropriate permissions.

Becker et. al.[22] propose an architecture to monetize resources on a Solid pod using blockchain transactions. According to them, each individual *Solid* pod would store the WebID and wallet address of the agent. An agent can be a seller or a buyer. The blockchain executes the payment process from the buyer, if the payment goes through, the seller is notified to add access rights to the buyer. However, they do not discuss the necessary means to model and apply the solution to capture AI-generated works of art.

**Semantic Technologies.** Bordoni et al. [138] developed an ontology prototype using Protégé to manage and record information about artists, artworks, and coloring materials. They represented a case study in which they created an ontology in order to build a repository for a subset of art. KGs suffer from incompleteness. So they always need to add some annotations to their KGs. Chen et al. [43] introduced a generative attitude to concentrate on minimizing the annotations for efficient knowledge discovery. Their model, KGGen, learns from the relational triplets in the knowledge graph without further annotations. They used a conditional multi-channel BERT network to model dependencies between head or tail entity and their relationship in the triplet regardless of entity pair co-occurrence in the text corpus.

## 4.5    Proposed Approach

To address RQ1.1, we present Platform for Independent Creators and Artists Supporting Self-Ownership (PICASSO), an architecture to manage the creation of works of art and preserve artists' interests. Figure 4.1 shows the main components of PICASSO. In the following, we describe how these components are integrated to achieve the main goal of preserving ownership.

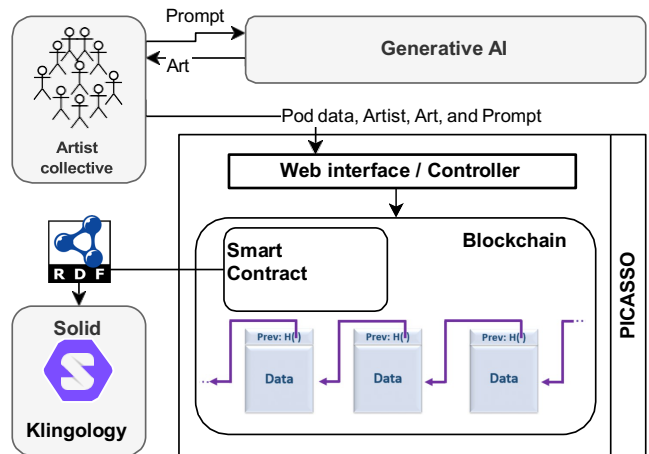

Figure 4.1: **PICASSO Architecture.** Decentralized technologies store and allow artists to commercialize their work.

**Web Interface / Controller.** This module is the main entry point of PI-CASSO. Artists start the process by choosing an operation. There are three main operations: (1) the inclusion of new art, (2) consulting open data about a specific work of art, and (3) a transaction to change the ownership of existing art. The Web interface checks for required parameters according to the requested operation. Parameters can include the digital version of the art, credentials to the personal *Solid* pod, the prompt used by the generative AI, and contextual knowledge about the artists involved.

After the request validation, the controller prepares the data, instantiates the necessary data structures, and continues the execution by registering the operation on the blockchain (case 1 or 3) or directly consulting art information on the Solid pod (case 2).

**Blockchain / Smart Contract.** As described in Section 4.4 with the Ethereum use-case, the combination of blockchain with smart contracts presents a suitable solution for copyright protection and trade. In our scenario, two operations will always be registered on the blockchain: the creation of a new piece of art, and the transference of ownership from one group of artists to another. Information stored on the blockchain is restricted to the identification of the entities involved in the transaction (e.g., artists) and the price of the artwork. Private information regarding the owner's consent is stored on their personal *Solid* data pod.

**Solid and RDF data modeling with Klingology.** To ensure the privacy of the art created using generative AI, the art with the metadata is stored in *Solid* as RDF triples and facilitates a decentralized environment for artists to protect their art and have a secure online marketplace for their art.

To answer RQ1.2, Figure 4.2 illustrates Klingology, an ontology to model the generated art prompt. We integrated existing ontologies the art ontology ArCo [34] with the ML ontology, i.e., MLOnto [29], and extended with additional concepts, for example, context about prompt purchase or ownership utilizing the Dublin Core vocab [20]. The underlying reason for the definition of ontology is to semantically ground the public subset of the data in the platform. This enables crucial aspects of semantic search and effective retrieval of prompts and related art pieces. The proposed framework is applicable in various marketplace applications, e.g., selling artwork as well as preserving the artist collective interests.

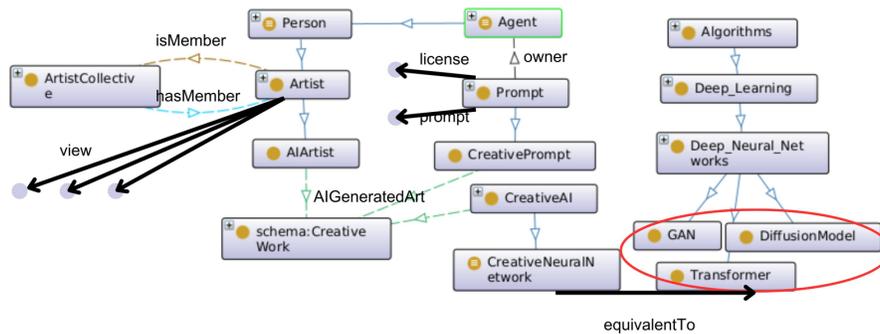

Figure 4.2: **Klingology**. Exemplar ontology depicts the extension of ARCO and MLOnto ontologies with the perspective of generated art prompts.

## 4.6  Theoretical Results: Use case

We demonstrate the feasibility of our approach through a description of a possible implementation and technical details about the workflow. As described in Section 4.5, the main actors and components involved are:

- Each *artist collective* member; who owns the art for the case of data insertion.

- An *artist* who wants to buy a prompt or a piece of generated art.

- A *Generative AI* system, owned by each artist collective and trained on their real art, in order to mimic and preserve the style of the artists. The specific implementations could be different from one artist collective to the other.

- *Web Interface/ Controller* component that facilitates the user with the proposed infrastructure implementing a ReST API or a complete web interface. This can be exposed through a web server of the artist collective.

- The *Solid* pod component is used to store and preserve RDF and non-RDF data regarding created prompts and pieces of art, as well as transaction records of each artist collective provided by users. The artist collective *Solid* pod can be linked to a specific account on a *Solid* provider or self-hosted[1]. The latter allows for greater data control. The *Solid* pod can be combined with the controller component to run the entire decentralized ecosystem.

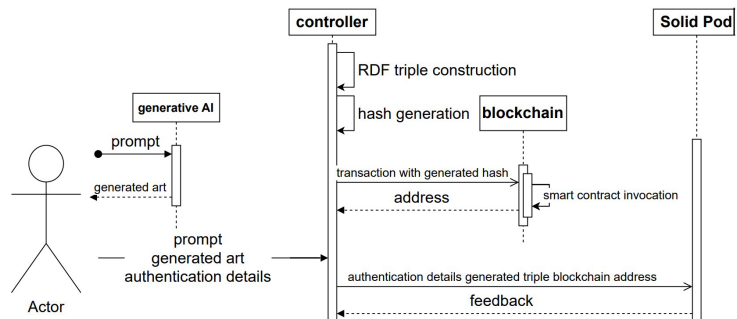

Figure 4.3: **PICASSO sequence workflow**

Figure 4.3 describes the workflow related to the implementation of the decentralized ecosystem, PICASSO, over generative AI. For instance, an artist collective wants to create a piece of music by specifying a prompt to their own generative AI system. The proposed framework sends both the prompt and the generated sound file with the WebID to the *controller* component. This ensures the validation of each artist collective stored over *Solid* pod. This interaction can be implemented by using an HTTP endpoint whose body of the request must contain the given prompt and the generated art. RDFLib [2] is utilized in the *controller* component to consider the ontology and create a knowledge graph that links the piece of art with the contextual knowledge of the artist (e.g., ownership) and license information based on Creative Commons [3].

The stored hash of the triple that links the generated data to the contextual information is sent as a payload of a transaction that invokes the smart contract for storing immutable data on the blockchain. The smart contract sends back the address generated by inserting the stored data into the chain. The next component involves communication with the *Solid* pod of the specific artist collective that matches the prompt given by the user. One of the most robust and well-used to interact with a *Solid* pod through its HTTP endpoints is Comunica [4]. For access control, authentication details are needed to be passed in the request, as well as the generated RDF triple and the transactional record of the blockchain. Lastly, the feedback from *Solid* pod provides the metadata of the stored RDF data. This feedback is stored in the blockchain as well to keep track of the successful insertion of data into the *Solid* pod. This way, the *Solid* pod offers protection over the access to the data, by defining ODRL [126] rules and restrictions so that only authorized users can access the data. General copyright protection is guaranteed by the ODRL rules of the associated Creative Commons license linked to the user's prompt and the generated piece of art. Blockchain allows immutably defining ownership of the prompt and generated art.

Whenever an external user wants to purchase a specific prompt about a piece of generated art, it first sends a query to the *controller*. In turn, it sends it to the *Solid* pod of the for accessing the resources and get the desired content. This interaction is tracked by submitting the query and the identifier of the user to the blockchain. Items are exchanged according to their prices. The token exchanged over a blockchain can refer to a specific coin created by the artist collective. The user sends to the *controller* all the information of the chosen art, including the Creative Commons licensing, used to check if the user's intention to use the art is compatible with the rules of the associated license. The controller modifies the ODRL rules associated with the chosen art, for instance, the generated sound file memorized in the triple stored in the artist collective *Solid* pod. To track data regarding the art, the exchange is regulated by creating a transaction that invokes a smart contract that manages to exchange the tokens from the user to the artists' collective wallet. At this point, the user receives the requested item, together with the related copyright and original owners metadata into its own *Solid* pod.

## 4.7    Conclusion and Future Work

In contrast to their efficiency, generative AI systems are becoming more and more detrimental for artists and creators regarding the copyright and ownership of their AI-generated collectives. We present PICASSO, a framework that utilizes decentralized technologies and semantic web technologies to protect the copyrights and ownership of AI-generated pieces of art. We used *Solid* to store the AI-generated artifacts' data and Blockchain technology to record and confirm transactions among the artists. The novelty of our work is that we guarantee a safe, trustable environment for the artists and owners of AI-generated artifacts using semantic web technologies.

For future works, it is necessary to discuss what is public and private art metadata, as well as discuss ownership regulations. Furthermore, identities of artist collectives and artists, in general, can be managed by using the Decentralized Identifiers (DIDs) [55], allowing the creation of digital identities that are decentralized and globally resolvable through a blockchain infrastructure.

**Acknowledgements**

We would like to express special thanks to our tutor Prof. Dr. John Domingue for his guidance and support. Also, we would like to show our gratitude to our distinguished faculty Prof. Dr. Harald Sack and Prof. Dr. Enrico Motta for their valuable contribution to the research study.

# Chapter 5

# ChatGPT, is this legal? Impact of the unintended misuse of AI solutions

IOANNIS DASOULAS, GEORGE HANNAH, MIKAEL LINDEKRANS, RITA SOUSA, MARY ANN TAN, CAITLIN WOODS, CLAUDIA D'AMATO

## 5.1  Semantic Web and Creative AI

Creative AI is a sub-field of AI aiming to generate artificial relics using existing digital content by examining training examples, learning their patterns and distribution [3]. Users can interact with creative AI tools by providing a prompt and receiving an answer that can have the form of a text, image, audio, video or some combination of the above. As with most new technologies, these tools are not immune to misuse, whether intended [167, 81] or unintended. While creative AI tools possess a broad coverage of information and have seen incredible success, the generated content can often be inappropriate for certain audiences or recommend actions that may be have legal implications.

Semantic web technologies and knowledge graphs (KGs) can play a significant role in addressing some of the issues posed by creative AI models. By incorporating KGs to creative AI tools, the models can access contextual information, enabling them to validate the facts they generate, be more coherent and filter their content better, providing additional information to the users.

## 5.2 Introduction

ChatGPT[1] and other Large Language Models (LLMs) are rapidly altering the way that we search for information. With proficiency at-par or even exceeding humans, these tools have become ubiquitous in all aspects of life, from education to software engineering [120]. LLMs are frequently being used for providing recommendations to users. However, research has suggested that users are inclined to blindly trust artificial intelligence (AI) tools, even when they are wrong [95]. The concern that we address in this report is that users may not be aware of the legal implications of the recommendations provided by the LLMs. In addition, a recommendation made in one country, may have different legal implications when performed in another country.

If a user asks a question to ChatGPT such as "how do I home-brew my own gin?", the OpenAI model (at the time of writing this paper) will provide a method and suggest that the user checks the local regulations for alcohol production. However, it will not go beyond this in assisting the user. In this report, we consider the possibility of presenting the ChatGPT users with *specific* laws that can potentially be violated when the AI's reply is acted upon by the user.

We propose a preliminary framework designed to fetch the legal context for answers provided by ChatGPT. This proposed framework suggests leveraging semantic web technologies and LLMs to enhance the knowledge presented to the user with additional legal information. We use the issues uncovered in developing this framework to identify potential future work and societal impact of this research avenue.

The remainder of this report is structured as follows: In Section 5.3, we give the research question that we consider in this report and Section 5.4 contains related work. In Section 5.5 we describe our conceptual framework for solving this problem and discuss how this framework could be evaluated in Section 5.8. Finally, in Section 5.9 we give a discussion of the opportunities and limitations of the proposed approach. We also comment on the potential societal impacts of this work.

## 5.3 Research Questions

Through the proposed framework in this report, we aim to address the following main research question:

*Would it be possible to interleave existing deep learning models with a legal KG in helping the user to understand the legal implication of their action?*

To answer the main question, the following sub-questions have been formulated.

A. What are the difficulties, if any, in modeling legal knowledge as KGs?

---

[1]ChatGPT, https://openai.com/blog/chatgpt

B. What is the general framework for interleaving legal KG and creative AI solutions?

C. How do we extract an answer and how is it presented to the users?

D. How do we measure the effectiveness of such a framework?

E. How can the societal impact of a framework, such as the one presented in this report, be evaluated?

## 5.4   Related Work

In order to represent legal information in a structured and uniform way, we have identified several KGs that represent legal knowledge. The European Legislation Identifier (ELI) [68] is a system to make legislation available online in a standardised format, so that it can be accessed, exchanged and reused across borders[2]. The Lynx [90] KG represents a collection of heterogeneous multilingual legal documents related to compliance legislation, regulations, policies, and standards from multiple jurisdictions. Lynx extended ELI (Lynx-LKG), used the EuroVoc thesaurus[3] and is available in RDF format. It is FAIR compliant and provides a SPARQL Endpoint. The Austrian Legal KG [66] represents legislative and judicial documents from Austria using EuroVoc and also provided a SPARQL Endpoint[94]. Although, these KGs manage to organize legal information, they do not provide related keywords in the form of metadata, for external systems to be able to identify laws in a more efficient way.

Over the last decade, KGs have been used for supporting QA systems [65]. Newer approaches that rely on LMs for question-answering, use retrieved documents to reinforce their answering efficiency [92, 98, 184, 119]. Retrieved information can be used for providing extra references to the user, as well as for fine-tuning the language models. However, current language model extensions that are fine-tuned to provide hyperlinks are unable to provide legal references[4].

## 5.5   Proposed Framework

The goal of this work is to provide relevant legal citations to the user when their queries have potential legal implications. We propose a preliminary framework for interpreting an answer from ChatGPT and augmenting that answer with the citation of a legal document. The proposed framework is summarised in Figure 5.1. When a question is asked to ChatGPT, a process is commenced to assess possible legal implications identified by the ChatGPT reply. First, the ChatGPT reply is assessed to see if it is relevant to some legal implication. A *Topic Modelling* approach can be used to identify if this is the case. As a general purpose chat-bot, it would be sufficient to use a domain agnostic model such as BERTopic [74]. If this is the case, the process enters a *Keyword Identification* phase. In this phase *Prompt Re-engineering* is first performed to get a more concise reply from ChatGPT. Following this, a *Keyword Processing* step disambiguates the terms so that they linked with the keywords in the Legal KG. This KG can then be queried to retrieve the relevant law (and a link to its citation) using the keywords in a *Answer Delivery* step.

---

[2]https://op.europa.eu/en/web/eu-vocabularies/eli
[3]EuroVoc, https://data.europa.eu/data/datasets/eurovoc
[4]https://chrome.google.com/webstore/detail/webchatgpt-chatgpt-with-i/lpfemeioodjbpieminkklglpmhlngfcn

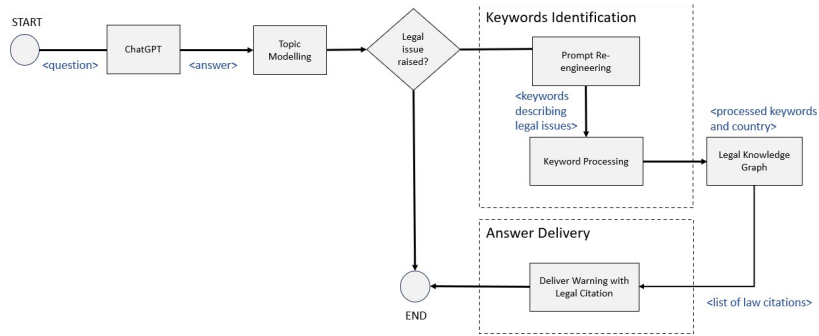

Figure 5.1: Framework for law recognition in answers from ChatGPT

The first step of the methodology is to construct a legal knowledge graph that can be used in identifying the relevant law from text. ELI ontology is used to provide the structure of the graph but, as mentioned in Section 4, this graph does does not contain information about the law at the text-level. Therefore, the ELI ontology is extended to include keywords that can be matched with the example from ChatGPT. To populate the knowledge graph with keywords, an entity recognition exercise could be performed over legal texts to extract *offences*, *actions*, *objects* and *people*. Based on our initial understanding of law texts and GPT answers, we believe that these categories are a good starting point for matching laws. To do this, we can make use of named entity recognition models for extracting knowledge from legal texts [41], or other domain-specific documentation [58].

We acknowledge that semantics are lost by storing keywords in the KG using a has_keyword relation. Future work could also involve defining *offences*, etc. as classes. An example of how this graph can be implemented in a use case,

After mapping the keywords in the answer and the legal KG, The third task includes using the legal KG to extract the laws that should be included in the answer and shown to the user. To do so, we generate structured queries, such as SPARQL, that will go from the keywords and return the corresponding KG entities representing legal resource subdivisions and corresponding URL addresses. Extracting triples from a KG using structured queries involves formulating queries that consider the structure and schema of our KG, i.e., the relationships between keywords and the legal resource identifiers. Once we have formulated the query, it can be executed over the KG using, for example, a SPARQL endpoint. The query results in RDF format can then be parsed to extract the specific information we want to show the user. Since multiple legal resources can be related to the same keywords, the query results will be ranked to select the most relevant legal resources related to more keywords. This list will then be delivered back to the user, along with the response from ChatGPT.

## 5.6 Resources

In this methodology, we make use of legal documentation and the ELI ontology schema. Today, laws for many countries are readily available online under relatively permissive usage licenses [165]. These laws are generally documented as large documents, separated into several subsections. Each of these subsections contain several expressions. Of course, these laws are complex. They contain legal jargon and contextual information. It is difficult to interpret this information with respect to everyday speech. Thus, when extracting keywords from this text in our methodology, we use domain-specific NER models that can handle technical jargon. In the proposed framework, we require structured knowledge describing the laws, their identifiers, the administrative region where the law applies. For this, we propose re-use of the ELI ontology since it allows access to law resources through structured and flexible identifiers. Additionally, we use a Legal Resource (and its parts), the Legal Expression that contains the keywords required for the pipeline, and the area where the law is applicable.

## 5.7 Use Case

In this section we describe an illustrative example of the proposed framework. We use an example legal expression from The Road Vehicles (Construction and Use) Regulations from the United Kingdom. The expression that we consider is Section : *"Subject to paragraph (4), every vehicle shall be constructed so as not to emit any avoidable smoke or avoidable visible vapour."* [166].

The model shown in Figure 5.2 describes one legal expression from the United Kingdom Government Legislation [166]. The expression describes an offence where a road vehicle has been modified and excessive and unnecessary emissions. The figure shows that the law itself is some eli:Expression that consolidates some eli:Legal Resource (i.e. The Road Vehicles (Construction and Use) Regulations 1986 part 61). The administrative area that the eli:Expression is relevant to is the UK.

In the example show in Figure 5.2, we also show some keywords that the proposed framework could use to identify a relevant law. For this illustrative use case, these keywords were extracted manually. However, in an implementation, the methodology described in Section 5 should be used.

The methodology can be tested with the prompt: "Given that I am in the United Kingdom, how can I change the muffler on my car's exhaust?". At the time of writing this paper, response that asks users to check the local regulations. From here, a engineered prompt can be used to ask ChatGPT for some clarifi-cation about the reasons for the legal issue and ask it to give some keywords. An example prompt is: "Regarding my question 'How do I remove my car's muffler?', please provide keywords of the relevant laws associated with the text: original answer". The keywords that are returned from this prompt are laws, regulations, emission standards, vehicle safety, vehicle performance, modified vehicles, exhaust systems, mufflers,
noise pollution, air pollution, emissions, fuel efficiency,
engine damage, professional mechanic, exhaust specialist.

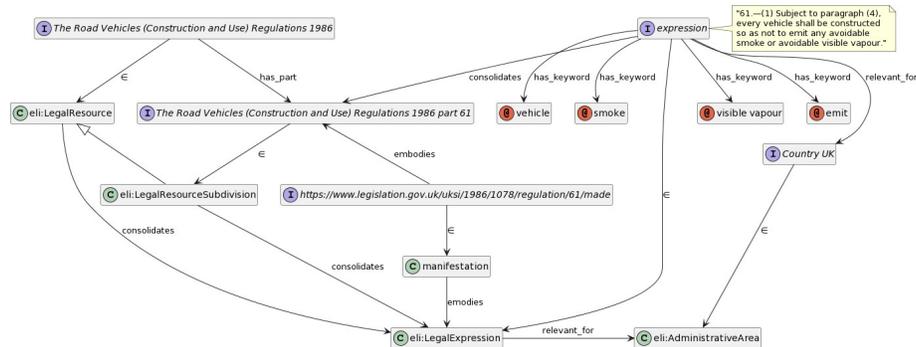

Figure 5.2: Legal KG for the Road Vehicles Regulations Example ("C" stands for "class", "I" stands for individual, and "@" is a string value)

The law can then be extracted using the approach described in Section 5. For example, the data to be returned to the user could be 'The Road Vehicles (Construction and Use) Regulations 1986 part 61" available at https://www.legislation.gov.uk/uksi/1986/1078/regulation/61/made.

## 5.8   Proposed Evaluation

Our evaluation comprises not only an overall evaluation of the framework but as well as its effectiveness in each individual task.

When considering the complete framework, the key part is the selection of a law, to deliver to the user. To evaluate this, we will require the input of a domain expert that can assess the performance of the pipeline by submitting prompts and providing judgement on the correctness of the responses returned by the framework.

**Legal Knowledge Graph Evaluation**

The quality of a KG is usually evaluated according to two aspects, consistency and completeness. First, consistency can be characterized by the lack of contradictions and can be validated through the process of reasoning using a reasoner, e.g. HermiT [71]. In future work, competency questions should be developed to validate the completeness of the legal KG in the context of law identification.

**Legal Keywords Identification Evaluation**

To assess the robustness of our approach, our framework should be applied to benchmark data-sets from different domains, which are publicly available and described in [155]. Performance metrics, such as precision, accuracy, and f-measure, can then be assessed. This evaluation will also allow comparison with existing entity-linking data sets.

**Answer Delivery Evaluation**

To evaluate the usefulness of our framework to users, we will conduct user studies [135, 99]. Participants should be selected from a population and presented with several questions asked to ChatGPT. Following that, a usability assessment such as the System Usability Scale [21] survey should be applied to understand the impacts of the designed answer. In addition, a higher-level impact study should be performed to evaluate the societal impact of the solution. For this, a good approach would be to perform qualitative interviews or focus groups, to gain a deeper understanding of usage motivations and what users plan to do with the increased accessibility of legal knowledge.

## 5.9   Conclusion and Future Work

We have shown that the proposed framework can complement the answers of ChatGPT by providing legal insight to the user. Given the limited legal context provided by ChatGPT today, we can look to further support user understanding by injecting related legal knowledge from a KG to the answers provided by ChatGPT. With respect to the difficulties in modelling legal knowledge as a KG (RQ1), there are existing legal KGs and ontologies, as well as their extensions that can be used off-the-shelf. It is important to consider that there might be legal implications for the users and groups looking to implement this framework, if a user acts on the instructions or answers provided by the AI, i.e. for advice-seeking questions. NLP tools such as topic modeling and keyword extraction can complement the pattern-based reasoning of SPARQL queries applied on KGs (RQ2). The evaluation of the effectiveness of the proposed framework is closely related to the domain knowledge encoded in the KG used to extract the appropriate law (RQ4). Components of the proposed framework will require a mixture of automated and manual assessment. Since the goal is to mitigate legal consequences for all parties, legal experts will provide the final verification. There are several potential societal impacts related to the use of LLMs, in this report we touch upon some of these in relation to promoting awareness and cautious trust, but are not aiming at providing a complete list. The most obvious impact that we see relates to the fact that interleaving LLMs with KGs would raise the legal awareness for users, which in turn promotes social accountability. Today, users are presented with general legal disclaimers but never presented with references to actual legal advice. Even in the case that the legal advice would turn out to not be exact or even wrong, one could argue that it raises awareness on the fact that you should not blindly trust the answers provided by LLMs (RQ5).

### 5.9.1 Future Work

The domain ontology (ELI) included in the proposal will be extended to consider additional contexts, such as agent, role, time and place, for a more precise result. The use case of the framework we propose is set in the UK.

The implication of including the different senses of the keywords needs further improvement. Adding a word-sense disambiguation component could be beneficial, in that it could increase the recall of the answer delivery component.

Laws and regulations go through a rigorous process before they are passed. However, they can be amended, and these amendments will require representation in the KG. Depending on the size and granularity of the KG, this comes with a certain complexity that we are not addressing in this report but identify as a key area for future work as the implications are that users are provided with outdated legal advice, or none at all. In this report we have proposed a basic design that fits for the completion of the use case (RQ3). A user interface can be applied to this framework to provide a complete service.

# Chapter 6

# KUBA: Knowledge Graph Generation through User-Based Approach

AVA SAHBI, CRISTIAN SANTINI, HASSAN HUSSEIN,
KERSTIN NEUBERT, VINCENZO DE LEO, XUEMIN DUAN,
IRENE CELINO

## 6.1 Introduction

The development of Knowledge Graphs (KGs) through the transformation of unstructured text has emerged as a highly promising research domain. KGs organize information in a structured manner, connecting real-world entities through semantic relationships. These graphs enable efficient knowledge representation, allowing machines to reason and answer complex questions. However, extracting structured data from unstructured text presents challenges due to the nature and heterogeneity of the data. Natural language processing and semantic understanding are significant obstacles in this process.

Traditional approaches for generating KGs from text, a task known as Knowledge Extraction (KE), rely on information mining and natural language processing techniques. These methods use machine learning algorithms to identify entities and relationships within texts and construct the knowledge graph. However, these approaches require rich human supervision and the development of detailed rules for each application. The manual extraction of information is time-consuming, labor-intensive, and prone to errors. Moreover, these approaches struggle with the heterogeneity and complexity of unstructured text, limiting their scalability and accuracy.

The emergence of Large Language Models (LLMs), such as BERT or OpenAI's GPT-3, presents new opportunities for generating KGs from unstructured text. These models, trained on extensive data, can learn complex semantic representations and autonomously generate Resource Description Framework (RDF) triples. This automated approach offers a more efficient and accurate method for KG generation. However, using LLMs for this purpose entails certain risks, such as hallucination or incomplete training data. Therefore, it is crucial to address transparency, limitations, and control over the outputs of the generation models.

This paper is structured as follows: Section 6.2 presents the scope of this work and our contribution. Section 6.3 discusses related work on Knowledge Extraction and LLMs. Section 6.5 presents the data used in the study. Section 6.6 explores various approaches to integrating Generative LLMs with human agents and automatic methods for KG extraction from text. Section 6.7 presents a potential evaluation of the proposed approach, and Section 6.8 concludes the paper.

## 6.2   Semantic Web for Creative AI

Creativity has a major role in Knowledge Engineering tasks for the Semantic Web, where human agents have to define consistent and complete formalization of a domain in a machine-readable manner. Scope of this research is to explore how recent Generative AI approaches for NLP/NLU may be used, along with humans and further automatic methods, to enhance KE processes and support humans in these tasks, by either reducing cognitive load or improving the quality of the generated output. The contribution of this work is the design of a framework for easily connecting Human Agents with Neuro-Symbolic methods to define and systematize hybrid pipelines. By investigating the effectiveness and usability of this framework, the research aims to contribute to the advancement of KG generation methods and the practical utilization of Generative AI in the semantic web domain.

## 6.3   Related Work

Prompt engineering and the use of large language models have garnered significant attention in recent research on Natural Language Processing (NLP) and Natural Language Understanding (NLU). Large language models like OpenAI's GPT and Google's BERT have revolutionized NLP tasks by leveraging their pretraining on vast amounts of text data and subsequent fine-tuning for specific downstream tasks. The concept of unsupervised learning through transformer-based language models was introduced by [137], which laid the foundation for subsequent advancements in prompt engineering.

Prompt engineering involves designing specialized instructions or queries to guide large language models in generating desired outputs. This technique has been extensively explored to enhance the controllability and performance of language models. [30] introduced the use of human-generated prompts to influence language model behavior and demonstrated its efficacy in various tasks. [104] investigated the applicability of prompt engineering for zero or few-shot learning in NLP/NLU tasks like text prediction or classification.

Several studies have analyzed the capability of Generative AI models, such as ChatGPT, to generate knowledge graphs. [136] empirically examined the zero-shot learning ability of ChatGPT and found that it performs well on reasoning-oriented tasks but faces challenges in specific tasks like sequence tagging. [70] demonstrated that ChatGPT outperforms crowd-workers in various annotation tasks, including relevance, stance, topic, and frame detection. Additionally, [161] evaluated ChatGPT's performance in extracting Knowledge Graphs from texts. The authors observed that including domain information in prompts, such as categories found in the input text, improves the quality of the resulting KG. However, the study also revealed that task-specific models outperform ChatGPT in this regard.

However, it is still to be clarified how Generative AI models can support Hybrid Human-AI systems for generating KGs out of structured and unstructured resources. More specifically, to which extent LLMs can fully substitute Knowledge Engineers in the creation of RDF triples from texts, and, if not, how can they be integrated in already available pipelines.

## 6.4 Research Questions

LLMs and the Semantic Web intersect in knowledge representation and extraction. LLMs excel in generating human-like text and aiding in natural language processing tasks, while the Semantic Web focuses on enhancing data organization, linking, and sharing on the web. LLMs have the potential to contribute to the Semantic Web by extracting and organizing information into structured formats, such as knowledge graphs or ontologies, improving the efficiency and accuracy of data representation. Based on these considerations, the following research questions are formulated:

- **RQ1:** How can we combine users and LLMs for extracting from given texts knowledge graphs that contain well-structured and reliable data?

- **RQ2:** How can we use Generative AI to generate KGs from text, allowing users to refine or create new prompts?

- **RQ3:** How can we involve the user with the minimum cost in the process of knowledge graph generation in collaboration with LLMs?

## 6.5 Resources

Exploring different configurations in which a human agent can be placed in a hybrid system requires having available different datasets that allow evaluating the response of the system in different cases. For these reasons we have prepared a first benchmark dataset [102] with the data already known to the model containing Wikipedia page abstracts for the representative pages of some categories of basic Wikipedia concepts, namely *Geography*, *History*, *Literature*, *Science*, *Technology*, *Mass Media*, *Sports*, *Politics*, *Natural Environment* and *Religion*

A second dataset [101], with data that the model has potentially never seen, was extracted from news published after the model's pre-training date. The news dataset has been extracted from the New York Times newspaper. Recent news stories have been specifically selected in such a way that they deal with recent discoveries or events that certainly contain information on specific entities that could not have been used at the time of its training. This dataset is public and available on Zenodo. The Large Language Model that we decide to use in our experiment is the gpt-3.5-turbo model provided by OpenAI.

## 6.6    Proposed approach

In the realm of Knowledge Graph construction, LLM such as gpt-3.5-turbo show the ability to extract RDF triples from textual data. However, a notable challenge arises that the vocabulary is self-generated when applying LLM to process the latest data that is absent from the LLM's training data and existing Knowledge Base, leading to potential complexities in KG integration due to LLMs autonomously designing the ontology. Compounding the issue is the practical constraint imposed by the maximum token limit in input, making it infeasible to present the entirety of textual data to the LLM in one prompt. In the absence of accurate and specific guidance during the individual feeding of articles, the resulting sub-graphs may exhibit substantial disparities, impeding integration with other components of the knowledge graph. This section discusses a new framework, namely Knowledge graph generation User-Based Approach (KUBA), which is independent from the underlying LLMs proposed and is designed to overcome the aforementioned challenges.

The KUBA framework is described by using compositional design patterns introduced in [77] to explain and systematize hybrid Neuro-Symbolic AI approaches. We extended this boxology by adding to the algorithmic components (ovals) Human Agents (HA) which act as decision makers in data integration steps. This framework comprises four basic variations which are described hereafter (Fig.6.2):

A **Zero-shot Triple Generation (ZTG)**: In this step, a prompt from the given text, containing task explanation and schema templates is created as in Figure 6.1. The population of the RDF schema template may come, for example, from user-defined SHACL shapes or Ontology Design Patterns and may not require full-fledged ontology specifications. Once the prompt is specified, the LLM (ML in the boxology) generates RDF triples based on the created prompt. This step entails leveraging user-defined schemata as guiding structures during the prompting process, aiming to enhance the coherence of the generated triples. By integrating domain-specific knowledge and explicit schema guidance, this method aspires to mitigate hallucinations and improve the fidelity of the LLM-generated RDF triples for the purpose of knowledge graph construction.

B **ZTG + Automatic Statistical Refinement:** To assess the semantic validity of the generated triples, this extension of the framework make use of a Machine Learning model employed to predict or classify valid triples out of the zero-shot generated KG. This can be executed, for example, by using a LLM which analyses the user-provided triples and aligns them with the semantic structure of the input text.

C **ZTG + Automatic Reasoning Refinement:** User-provided schemata or Shapes Constraint Language (SHACL) [96] shapes can be provided in order to automatically filter out triples which do not semantically conform to the requirements of a given Ontology using a Knowledge Reasoning system.

D **ZTG + Human Refinement:** Human Agents directly assess the quality of the generated KG by refining or enhancing the output.

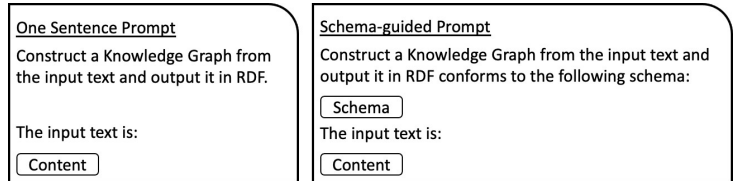

Figure 6.1: A template for baseline prompt and schema-guided prompt.

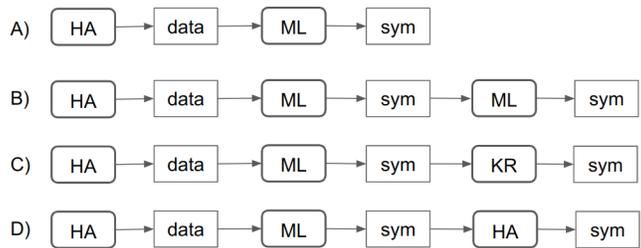

Figure 6.2: Schema of different approaches proposed. A) is the baseline approach. In proposal B) we added another step of a ML model, in proposal C) we substituted the final ML model with a Knowledge Reasoner and in the last proposal D) we substituted the ML or the KR with a Human Agent.

e

## 6.7   Evaluation of approaches

We propose to evaluate the suggested approaches using two datasets as inputs: (1) Wikipedia page abstracts and (2) New York Times News (section 6.7.1) and a set of metrics described in 6.7.3 for comparison. We intend to involve users that have already experience in knowledge graph generation, who can judge the complexity of the task and quality of the result. We access the required human costs in the human-AI interaction by measuring the effort from a user-perspective with a questionnaire that asks for (1) Time consumption, (2) Difficulty level, (3) Satisfaction with the result.

### 6.7.1   Resources

Exploring different configurations in which a human agent can be placed in a hybrid system requires having available different datasets that allow evaluating the response of the system in different cases. For these reasons we have prepared a first benchmark dataset [102] with the data already known to the model containing Wikipedia page abstracts for the representative pages of some categories of basic Wikipedia concepts, namely *Geography*, *History*, *Literature*, *Science*, *Technology*, *Mass Media*, *Sports*, *Politics*, *Natural Environment* and *Religion*

A second dataset [101], with data that the model has potentially never seen, was extracted from news published after the model's pre-training date. The news dataset has been extracted from the New York Times newspaper. Recent news stories have been specifically selected in such a way that they deal with recent discoveries or events that certainly contain information on specific entities that could not have been used at the time of its training. This dataset is public and available on Zenodo. The Large Language Model that we decide to use in our experiment is the gpt-3.5-turbo model provided by OpenAI.

### 6.7.2   Baseline for comparison

We utilize gpt-3.5-turbo, a Large Language Model trained on extensive textual data, as the baseline for our experimental evaluations. With a maximum token limit of 4,096, gpt-3.5-turbo is capable of generating coherent and contextually relevant responses when provided with a prompt. In our experiments, we employ one-sentence prompting as the baseline approach. The prompt used in our experiments is depicted in Figure 6.1 for reference.

### 6.7.3   Evaluation metrics

After KG construction with, the main characteristics of the Knowledge graph are measured by basic statistics on RDF triples and instances. The type of qualitative evaluation criteria for a knowledge graph highly depends on the application e.g. semantic search, decision making or knowledge management and the available resources e.g. ontology. Therefore we propose that the human agent selects upon a list of six general KG quality metrics (from *Wang et al. 2021* [171]) and six structural quality metrics (according to *Seo et al. 2022* [152]). Further quantitative evaluation relies on the availability of complete ground truth data i.e. a knowledge base(s) that cover all triples, which is usually not available. Some criteria like Trustworthiness and Availability are subjective and can be only evaluated manually by the human agent. At the end of the human-AI interaction an outcome evaluation is produced that summarizes the performance of the approach according to the user-selected evaluation criteria. The human agent decides the validity of the produced knowledge base and in case of rejection, further knowledge graph refinement might be done by a ML model (approach B), a Knowledge Reasoner (approach C), or a human agent (approach D).

**Basic metrics**

A. Number of triples in RDF data

B. Number of instances in RDF triple data

C. Number of classes for specified ontology

D. Number of properties for specified ontology

**KG quality metrics**

Wang et al. 2021[171] propose the following six evaluation dimensions to assess KG quality:

A. **Accuracy:**

- Syntactic validity of triples
- Semantic accuracy of triples

B. **Consistency:** No contradiction of triples (logical/ formal)

C. **Completeness:** The degree of information coverage in a particular dataset.

- Population Completeness
- Interlining Completeness

D. **Timeliness:** Up-to-dateness of content.

E. **Trustworthiness:** Degree of acceptance of information as correct by the user.

F. **Availability:** Extend of availability of information to the user.

**Structural quality metrics**

A. **Instantiated Class Ratio (ICR):** refers to the ratio of instantiated classes relative to the total number of classes in the Ontology. ICR indicates the representation of ontology classes in the knowledge base.

B. **Instantiated Property Ratio (IPR):** refers to the number of properties used in RDF triples relative to all properties of the Ontology. IPR measures the coverage of properties of the ontology in the knowledge base.

C. **Class Instantiation (CI)**: For all subclasses the portion of the instances of the class to the total number of instances is calculated. The sum is normalized by a penality term according to the depth of the subclass. CI is a measure for the division of classes into subclasses.

D. **Subclass Property Acquisition (SPA)**: A function is applied to the difference of the property sets in subclass and superclass relationships. The obtained number of properties is summed and normalized by the number of classes. SPA measures the number of properties in the subclass that are not contained in the superclass in the ontology.

E. **Subclass Property Instantiation (SPI)**: SPI measures unique properties of the subclasses in the RDF triples that are not contained in the superclass.

F. **Inverse Multiple Inheritance(IMI)**: The inverse of the average number of superclasses per class is computed to obtain the average multiple inheritance. IMI measures the simplicity of the knowledge graph, which makes complex queries more feasible.

## 6.8    Discussion and Conclusions

This paper presented a Hybrid Human-AI framework to design simple architecture to integrate Generative LLMs, ML, KR and User Agents in the process of automatically creating KGs from text. This work specifies the functionalities of different components in the Knowledge Extraction pipeline and how humans, e.g. domain experts and knowledge engineers, may be involved in the pipeline, by specifying ontology requirements in prompts, refinement strategies and system preferences.

However, LLMs still show significant limitations in the capabilities of extracting relations between concepts and real-world entities which are semantically contained in the input text. Issues related to bias or incompleteness in training dataset hinder the applicability and scalability of these approaches. The limitations are also made worse by the current scarceness of Generative LLMs for the public domain.

However, this work does not give a specific answer to the question if current Generative LLMs are capable of not of extracting semantically and syntactically correct KGs from seen or unseen texts. In future work, the capabilities of LLMs of identifying valid entities and relations, wrt. to different prompt engineering strategies and refinement techniques, should and will be investigated. Moreover, a qualitative and quantitative evaluation of the obtained KGs from the extracted dataset will be presented, involving users with a variety of experiences and backgrounds.

**Part III**

# The Sound of Silence - Commonsense and Implicit Knowledge in Large Language Models

# Chapter 7

# Draw Me Like Your Triples: Leveraging Generative AI for the Completion of Wikidata

RAIA ABU AHMAD, MARTIN CRITELLI, SEFIKA EFEOGLU, ELEONORA MANCINI, CÉLIAN RINGWALD, XINYUE ZHANG, ALBERT MEROÑO

## 7.1    Research Questions

Our report addresses the following research questions (RQs):

- RQ1: To what extent can the output of generative AI be used for knowledge graph completion?

- RQ2: To what extent do prompts increasingly similar to natural language can be used in text-to image models to produce higher quality images?

- RQ3: To what extent do prompts increasingly similar to natural language can be used in text-to image models to preserve the emotion and sentiment between prompts and generated images?

## 7.2    Semantic Web for Creative AI

Our definition of *Semantic Web for creative AI* is circular. On the one hand, we think that Semantic Web technologies can be used for knowledge-aware prompt engineering to be used as input for generative AI. And on the other hand, generative AI can be a powerful tool for multimodal KG construction and completion. Our approach demonstrates this full circle by using triples of Wikidata items as prompts for a text-to-image model, and then using the creative AI output to complete missing images for the same items.

## 7.3 Introduction

Recent advances in artificial intelligence (AI) have given rise to automatic systems that can generate images from textual descriptions [185]. Thus, the field of *creative AI* has witnessed an increasing number of use cases and applications. Consequently, researchers have been looking into methodologies for enhancing the textual prompts given to creative AI models, in order to exploit its inner features and receive the best results [131].

Semantic Web technologies have been incorporating AI into many of their applications and tasks. Some examples include ontology creation and enhancement, knowledge graph (KG) embeddings, and KG completion (KGC) [79]. In this report, we will focus on the task of KGC, i.e. completing missing information in KGs, by leveraging creative AI applications and using different prompt engineering approaches.

To exemplify our idea, we focus on one aspect of KGC in Wikidata [170]: image completion of fictional characters. Querying Wikidata shows that only 7% out of the 83.7K instances of the fictional character class (including its sub-classes) have an image. To resolve this, we propose the following method: 1) Extracting knowledge about fictional character entities in Wikidata in the form of triples. 2) Creating different types of prompts to portray those triples. And 3) Using a text-to-image model to create images portraying the fictional characters.

We then present two strategies for evaluating our approach. The first consists of constructing a ground-truth dataset of fictional character entities that do have an image on Wikidata, feeding those to our pipeline, and comparing the resulting images to the ground-truth ones. The second approach compares the sentiment and emotion of each prompt to that of its corresponding image. This is based on our intuition that having similar affective features of the prompt describing the character and the generated image depicting it indicates that generative AI can indeed be used for the purpose of KGC, and more specifically for image completion of characters.

This project contributes to the ever-growing area of research that leverages AI technologies for KGs. To the best of our knowledge, no previous study has explored the realm of using generative text-to-image AI for KGC.

## 7.4 Related Work

**Generative AI:** Current groundbreaking advances in AI enable machines to generate novel and original content based on textual prompts. Such generative applications include text-to-text [83], text-to-image, and even text-to-music. Generally, these models can capture complex patterns from the input text and produce coherent outputs. A recent survey [185] shows that text-to-image ap-plications specifically have been emerging since 2015, when AlignDRAW [107] pioneered the field by leveraging recurrent neural networks (RNNs) to encode textual captions and produce corresponding images. Since then, end-to-end models started leveraging architectures such as deep convolutional generative adversarial networks (GANs) [141, 186, 181], autoregressive methods [139, 59, 179], and the current state-of-the-art diffusion-based methods [123, 144, 143].

**Prompt Engineering:** Because of the afore-mentioned advances, a novel area of *prompt-based engineering* has emerged, in which humans interact with AI in an iterative process to produce the best prompt (i.e. textual input) for a specific desired output [105]. Recent work has shed light on prompt engineering for AI art specifically [131], concluding that simple and intuitive prompts written by humans are not enough to get desired results. Rather, writing good prompts is a learned skill that is enhanced by the usage of specific prompt templates and modifiers [130]. Naturally, prompts themselves can also be automatically generated and do not necessarily require human manual writing. A novel research approach introduced mechanisms to enrich and refine prompts in order to create images with pre-defined emotion expressions [173]. This was done by extracting text features and applying strategies such as adding emotion words, context, paraphrasing, and increasing concreteness. However, to the best of our knowledge, no work has been done yet on leveraging Knowledge Graphs (KGs) for prompt engineering.

**Knowledge Graph Completion:** KGs are essential knowledge structures that have been increasingly used to solve many knowledge-aware tasks [79]. A large number of KGs have been constructed and are being used and edited by users such as DBpedia [100], Freebase [26], YAGO [159], and Wikidata [170]. However, a major problem of KGs remains their incompleteness, as many relations and entities are still missing [79]. As such, the task of KG completion (KGC) introduces many algorithms that can be leveraged for predicting missing parts of triples based on existing ones [154]. Researchers have leveraged some generative AI models to solve this task, however, existing studies use text-to-text generation, e.g., converting KGC to a sequence-to-sequence generation task using a pre-trained language model [180]. Another similar work tackled this problem with a similar solution by verbalizing graph structures and using flat text and using it as input for a language model [42]. In contrast to previous studies, our work proposes contributing to KGC by leveraging text-to-image generative models for missing images in Wikidata, using human fictional characters as a use-case.

## 7.5 Resources

All the code required to reproduce our results, including the evaluation framework, is openly available and can be accessed at https://github.com/helemanc/gryffindor.

**Dataset Creation:** We firstly began by designing a Wikidata SPARQL queries,

that allows us to obtain all the triples related to the fictional characters[1]. In regards of the limited time we have for conducting our study we sampled a dataset of 100 randomly selected entities that have images. This first sample allowed us to have a first intuition and a first basis for conducting our analysis. As second step we computed some statistics about theses triples : 100 different predicates are invested into our subsets and each entity are described by 9 different predicates. The top 10 predicates used are the following one : 'instance of' (99), 'sex or gender' (75), 'present in work' (71), 'occupation' (42), 'given name' (39), 'creator' (36), 'country of citizenship' (32), 'performer' (32), 'from narrative universe' (23), 'languages spoken, written or signed' (21) and 'spouse' (19). Please note that the predicate *altLabel* is discarded since only English is considered in our initial evaluation.

**Prompt Generation:** We designed 4 types of prompts, based on the data extracted from the knowledge graph, please refer to section 7.6.2 for more details. Among them, the verbalized triples are generated by the WDV verbalization model[2] proposed by Amaral *et al.* [8].

**Image Generation:** For the image generation task, we proceeded to apply DALL·E 2[3] and the free version of Craiyon v.3[4]. We decided to use two different AI models to compare them considering three main aspects: open-source availability, computational power, and output modalities. DALL·E 2 has been released in January 2022 and constitutes an improvement of DALL·E in text comprehension and image resolution. In addition to these elements, the second version allows users to load and modify their own images. In this work, it was applied via the official API that makes it possible to quickly generate one image per prompt. Craiyon is formerly a web-based mini DALL·E 2 created by Craiyon LLC. Able to generate a group of nine images from the same prompt according to different stylistic effects. In addition to this, this model offers users the possibility to exclude certain concepts from the image generation process using the "negative words" slot. For the image generation task, these two models are used and the results obtained analyzed and compared.

**Evaluation Framework:** To assess how the affective content of the prompt influences the representation of affective content in the generated image and determine whether the character from the original image remains recognizable, we employed two distinct models: Emotion English DistilRoBERTa-base[5] and DeepFace[153]. For inferring emotions from the four types of prompts, we utilized Emotion English DistilRoBERTa-base. This model, a distilled version of RoBERTa, was trained on a diverse range of datasets containing emotion-labeled texts from platforms like Twitter, Reddit, student self-reports, and TV dialogues. It focuses on predicting the six emotions defined in Ekman's emotion model[60] (*anger, disgust, fear, joy, neutral, sadness* ) along with *surprise*. We selected this model due to its alignment with the emotions considered in DeepFace and its demonstrated high performance. DeepFace is an agile Python framework designed for face recognition and comprehensive analysis of facial attributes. It seamlessly integrates several cutting-edge models, including VGG-Face, Google FaceNet, OpenFace, Facebook DeepFace, DeepID, ArcFace, Dlib, and Sface. In our study, we specifically harnessed the power of VGG-Face within the DeepFace framework. This allowed us to extract a wealth of valuable information from both the original and generated images, encompassing attributes such as age, gender, emotion, and race. Among this diverse array of attributes, we focused on leveraging the emotion feature for our research purposes.

---

[1]The package *qwikidata* (https://pypi.org/project/qwikidata/) is used for triple retrieval.
[2]https://github.com/gabrielmaia7/wdv
[3]https://openai.com/dall-e-2
[4]https://www.craiyon.com/
[5]https://huggingface.co/j-hartmann/emotion-english-distilroberta-base

## 7.6    Proposed approach

This section outlines a method for image generation corresponding to Wikidata entities currently lacking visual representation. Our primary focus for this report is the class designated as *fictional character*. The initial stage of our approach involves extraction of relevant triples pertaining to a specific entity. Subsequently, these triples are used to form various types of prompts, functioning as inputs to a sophisticated text-to-image artificial intelligence model. The ultimate goal is to generate suitable images that can serve as accurate visual identifiers for their corresponding Wikidata entities. This pipeline is shown in Fig. 7.1.

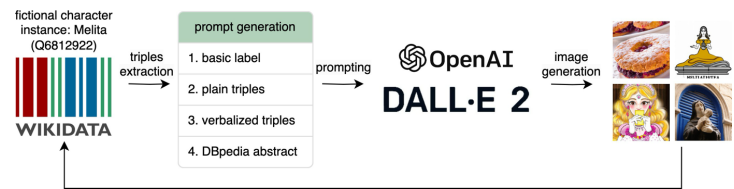

Figure 7.1: The pipeline of our proposed KG completion process.

### 7.6.1    Triple Extraction

Generating an image for a specific entity requires its related triples to define the characteristics. In Wikidata, these can be obtained through SPARQL queries, yielding all triples with the entity as the subject. Moreover, since properties and entities are represented by IDs, we also extract and translate each triple from IDs to their corresponding labels.

### 7.6.2    Prompt Generation

For each unique entity, we generate several distinct prompts. The first three prompts are created utilizing the set of triples that have been extracted for the respective entity. These prompts are described as follows:

A. *Basic Label* : This prompt merely employs the "label" that Wikidata assigns to its entities. For instance, given the entity Q692395, the corresponding basic prompt is denoted as *Habinnas*.

B. *Plain Triples*: This prompt is derived by concatenating the subject, predicate, and object of a triple to form a single coherent sentence, verbalising all available triples linked to a specific entity. For instance, for the triple [Habinnas, occupation, craftsman], the sentence created from this triple is given as ***Habinnas is an occupation craftsman***.

C. *Verbalised Triples*: Triple verbalization is defined as the transformation of structured data (i.e., triples) into human-readable formats (i.e., text). The verbalized triples can serve as a paragraph of summarization of input triples.

D. *Dbpedia Abstracts*: We also use Dbpedia abstracts as prompts, obtained by querying the English chapter of Dbpedia *et al.* [31]. Different from the above prompts, the abstract is edited by human editors.

Please note that for (2) and (3) above, the duplicate predicate of input triples except *instance of* is removed to avoid information duplication.

### 7.6.3   Image Generation

To run the text-image process, prompts were applied according to the distinction of the categories described above. First of all, Craiyon was applied without resorting "*negative words*" and selecting "*none*" for the style effect. Of the nine images generated, we opted for the one with the best resolution. DALL·E 2 has been used to apply the official API to automatically generate one image per prompt. In both generation processes, the output images were grouped by reference resource and sorted according to the resource ID provided by Wikidata. As Craiyon requires the manual insertion of prompts, we proceeded to select and organize a limited corpus of sample images to quantitatively and qualitatively compare the performances of two A.I. models.

### 7.6.4  Evaluation Framework

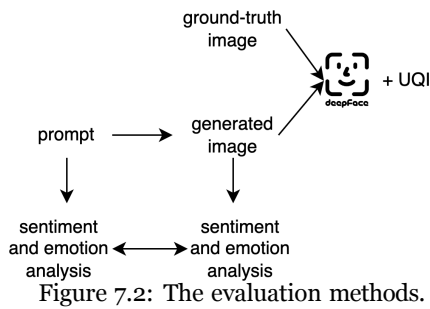

Figure 7.2: The evaluation methods.

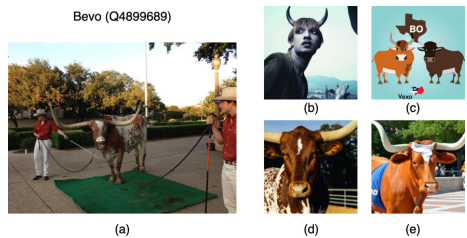

Figure 7.3: Generated images for *Bevo*. (7) Ground truth from Wikidata. (b) Basic label. (c) Plain triples. (d) Verbalised triples. (€) Dbpedia abstracts.

Two different evaluations are used to evaluate the characteristics of generated images, and how are they related to the types of prompts, as shown in Fig. 7.2.

A. The generated images are compared to the ground-truth image in Wikidata. For image similarity metrics, the Universal Quality Image Index (UQI) was utilized. UQI computes a pixel-based similarity score by comparing the generated images with their corresponding original images. Notably, since the majority of the original images are in grayscale, the similarity computation also takes into account their grayscale versions. UQI evaluates "image quality based on factors such as loss of correlation, luminance distortion, and contrast distortion" [168]. In a previous study [168], various pixel similarity metrics were compared, and the authors proposed UQI as an ideal metric. We have decided to leverage UQI metric for our evaluation distortion.

B. To assess the impact of prompt nature on generated image characteristics, we conducted a comparative analysis of prompts and resulting images. Our study specifically focused on evaluating the sentiment and emotion expressed in both elements, enabling a direct comparison. Emotion English DistilRoBERTa-base and DeepFace models were employed to predict emotions and extract information, respectively. The highest-scoring predicted emotion was considered for analysis, and sentiment analysis was performed for both prompts and image content. By following the widely accepted assumption[10] that certain emotions are perceived as positive or negative, we gained insights into the overall emotional tone conveyed by the prompts and images.

## 7.7  Results

**Emotional Analysis:** Table 7.1 shows the coherence of emotions and sentiments between prompts and their corresponding images. It investigates how well the emotions and sentiments expressed in the prompts align with those observed in the generated images, providing valuable insights into their consistency and correlation. The coherence score is based on matching emotions/sentiments between prompts and images. The results demonstrates that the highest coherence is achieved when the image generator is prompted with verbalized triples as input.

| Prompt Type | emotion | sentiment |
|---|---|---|
| Basic Label | 10 % | 53 % |
| Plain Triples | 34 % | 62 % |
| Verbalised Triples | 35 % | **86** % |
| Dbpedia Abstract | 42 % | 80 % |

Table 7.1: Coherence of Emotion and Sentiment between Prompts and Images

**Image Evaluation:** The DALLE-2 image generator has been used to generate images for 100 fictional characters across four prompt types. Table 7.2 shows the mean, minimum, and maximum UQI scores computed for the four different prompts. The basic label prompt exhibits the lowest mean, maximum, and minimum similarity scores compared to the other metrics. Conversely, the plain triples prompt yields the highest mean similarity scores, followed by the verbalised triples prompt. Consequently, relying solely on the basic label prompt is insufficient for generating images of fictional characters.

| Prompt Type | min | max | avg |
|---|---|---|---|
| Basic Label | 0.170945 | 0.792142 | 0.484021 |
| Plain Triples | 0.241029 | **0.848880** | **0.582811** |
| Verbalised Triples | **0.251560** | 0.674282 | 0.523799 |
| Dbpedia Abstract | 0.171420 | 0.766946 | 0.508574 |

Table 7.2: Comparison between generated and ground-truth images based on UQI

## 7.8    Discussion and Conclusions

Our findings demonstrate that enhancing prompts, through verbalization or inclusion of Dbpedia abstracts, significantly improves the coherence of emotions and sentiments between prompts and images. This highlights the potential of using natural language-like prompts in text-to-image models to effectively preserve emotional and sentiment aspects (addressing RQ2). Moreover, we cannot say if enriching prompts improves image quality or not, as the evaluation metric fails to capture semantic content adequately. Thus, we cannot answer RQ3.

To address the limitation of image-similarity evaluation, future experiments will incorporate alternative measures such as DICE metric[85]. Additionally, human evaluation is crucial to better understand how prompt content and composition influence image quality (RQ2 and RQ3). Furthermore, we plan to evaluate on a larger and more diverse dataset, including testing on a dataset without ground truth, relying solely on human evaluation. Finally, we aim to expand our study to other domains, such as fictional cities and abstract concepts, for a broader understanding of prompt influence.

## Acknowledgement

This project made use of the central High-Performance Computer system at Freie Universität Berlin, and we would like to express our gratitude for the resources provided. We would also like to extend our thanks to our tutor, Albert Meroño Peñuela, for generously providing us with his OpenAI API Key, but more importantly, for his invaluable input and help.

# Chapter 8

# A is the B of C: (Semi)-Automatic Creation of Vossian Antonomasias

GIADA D'IPPOLITO, NICOLAS LAZZARI, ANOUK M. OUDSHOORN, DISHA PUROHIT, ENSIVEH RAOUFI, JOHANNA ROCKSTROH, SEBASTIAN RUDOLPH

## 8.1   Research Questions

A. How does the quality of Vossian Antonomasias produced by a mixed-method approach of symbolic representation and embedding-based models compare to output generated by a large language model such as ChatGPT?

B. Are there any differences in the quality of the results produced by the knowledge graph embedding-based method and lexico-semantic embedding-based method?

C. Can the simple method based on TransE for knowledge graph embeddings be improved by a more suitable, elaborate vectorspace operation, better capturing the intuition behind VAs?

## 8.2   Semantic Web for Creative AI

Can algorithms or machines exhibit creativity when provided with sufficient human knowledge in symbolic and numerical form? This question raises interesting considerations. Vossian Antonomasias (VAs) can serve as a benchmark for exploring creativity in machines. By leveraging vast human knowledge, VAs demonstrate a degree of creativity in generating innovative solutions. Furthermore, the concept of Vossian Antonomasias combines semantic web resources and AI to create stylistic devices perceived as creative. This challenges our understanding of creativity and its potential in AI systems.

## 8.3  Introduction

Vossian Antonomasia (VA) is a popular stylistic device for describing one entity by referring to another, typically in a witty and resourceful manner. A VA expression consists of three parts: target (A), source (B), and modifier (C), composed into a statement of the form

<div align="center">A is the B of C.</div>

Therein, the target A is described using the combination of source and modifier. VAs are now a welcome element in many journalistic genres and frequently appear in headings, as they can be both informative and enigmatic, and are well-known for their unique entertaining qualities. In general, the source B is a well-known, widely recognized named entity from which one or more typical traits or characteristics are to be invoked. The target, on the other hand, does not have to be a specific entity. For instance, *"I was the Shakira of the school,"* and many more of these examples can be found here [1]. During her school days, the *source* entity refers to herself as the *target* Shakira in the preceding example.

For a human writer, a good knowledge of the specifics of the target entity is critical in the creative process of arriving at a VA expression. One has to identify a salient set of characteristics of A that is similarly, or even more prominently realized by B.

In this paper, we propose an approach to the task of automatically generating VAs by exploiting the latent semantic capabilities of vector space embeddings. A higher-level representation of our approach is expressed visually in Figure 8.1, which shows the process we followed in order to obtain our results. Firstly, in order to extract potential candidates for source B, we employ SPARQL queries over the Wikidata endpoint. After the extraction process, we leverage existing embeddings into vector spaces to find vector representations for A, and C, as well as all candidates for B, and use vector-based techniques to identify the (arguably) most suitable B among the candidates.

In order to explore the added value of Knowledge Graphs, we decided to try both Knowledge Graph embeddings and text-based embeddings. As for the techniques for identifying the most suitable B, we compared a basic approach based on TransE with a novel, more elaborate one, using projections.

---

[1] https://vossanto.weltliteratur.net/emnlp-ijcnlp2019/vossantos.html

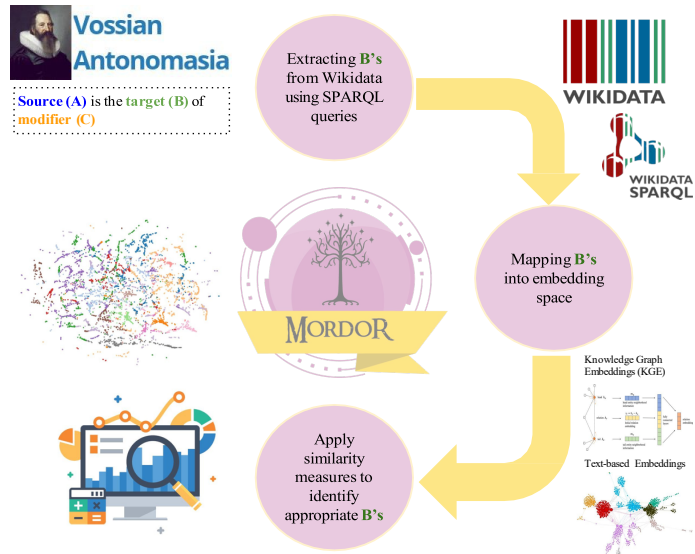

Figure 8.1: **Vossian Antonomasia (VA) pipeline:** The figure demonstrates the process that will be followed in the current research work.

## 8.4 Related Work

The authors of [67] demonstrated that by using Wikidata, they were able to overcome the shortcomings of available NER tools and confirmed that VA is a linguistic and cultural phenomenon. Through quantitative VA explorations, they were able to capture the phenomenon as a whole, encompassing the source, target, and, when available, modifier. Their approach involves searching for a network of individuals interconnected by diverse modifiers, where the nodes can function as either sources or targets. This network aids in understanding hidden patterns of role models, revealing how they vary across countries and languages. However, a limitation they acknowledged is their reliance on the most prevalent pattern of VA, namely, *"the...of"* which resulted in the omission of numerous expressions extracted from the New York Times (NYT) corpus. For instance, notable phrases such as *"the American Oscar Wilde"* and *"Harlem's Mozart"* were overlooked despite their significance. Another approach for the extraction of VAs is presented by [150]. They focus on the extraction of the target by using *coreference resolution* and visualize the connections between the source and target entities extracted in the VAs in form of a web demo. The authors of [149] used neural networks for the end-to-end detection of VAs resulting in two models: one for binary sentence classification of and another for sequence tagging of all parts of the VA on the word level.

Our approach, however, focuses on the generation of VAs with a given source and a number of targets to be selected from. Up to this point, this It further compares the results generated with Knowledge Graph Embeddings, Word Embeddings and and ChatGPT as a Large Language Model baselines. Similar to the approach of [67], out output will only focus on the pattern *"the...of"*.

TransE [188] is a knowledge-base embedding model which models relationships by interpreting them as translations of the entities' low-dimensional embeddings. In TransE relationships are represented as translations in the embedding space. If *h,r,t* is true, then the embedding of the tail entity *t* should be close to the embedding of the head entity *h* plus some vector determined by the relation *r*.

Based on the distributional hypothesis, word Embeddings [6] are vectors composed of fixed-length, dense, and distributed representations of words. Word embeddings have been discovered to be extremely useful for a variety of Natural Language Processing (NLP) tasks, including but not limited to chunking. GloVe [2]is an unsupervised learning algorithm for generating word vector representations. The representations are trained using aggregated global word-word co-occurrence statistics from a corpus, and the resulting representations highlight interesting linear substructures of the word vector space.

## 8.5   Resources

We used Wikidata, accessed via the SPARQL endpoint [3] to extract entities and domains corresponding to the source, target and modifier as a basis for our production pipeline. We chose the Wikidata Knowledge Graph because of two reasons: just like DBpedia, Wikidata contains a large number of triples [4] with a broad coverage of encyclopedic knowledge, which enables us to select a sufficient sample for every component of the Vossian Antonomasias. The second reason is that Wikimedia provides more structure and consistency when it comes to entities. Although DBpedia contains many more items than Wikidata *add source*, it does not store information, such as numbers or relationships, in a language-independent manner [2], which leads to duplicate entries.

## 8.6   Proposed Approach

At first, we sample the targets with the help of selection methods using SPARQL to obtain fictional entities in general and humans that correspond to the following domains:

|  |  |  |
|---|---|---|
| · sports | · computer science | · mathematics |
| · politics | · entrepeneurship | · genre (musicians & artists) |

Simply calling all humans and filtering afterwards gave us a timeout error which forced us to be more creative, leading to this approach. To ensure that the targets are popular enough, we restrict our search to entities that possess at least 31 translations for fictional characters and at least 70 for real-world individuals with the help of site links: the amount of languages in which there exists a Wikidata page for that specific target.

---

[2]https://nlp.stanford.edu/projects/glove/
[3]https://query.wikidata.org/
[4]According to Wikidata's statistics the knowledge graph currently contains 104,204,236 items. [177]

For the sampling process, we search Wikidata for entities that possess a certain amount of salient properties. This ensures that we are able to map varying subsets to the target and the modifier correspondingly. To generate sentences, we rely on knowledge graph embeddings using TransE [28] trained on the Wikidata-5M dataset [172] as well as word embedding methods using word2vec [110] and GloVe [132] as provided by *gensim* [142]. Given an A, we decided to take A's occupation as the C. In case there are multiple occupations given, we take the first one listed. The idea of an VA is that the B comes from a completely different field, but is still similar in a way. We rely on two different methods: a translational based method and a projection based method. The translational based method follows the intuition of *TransE* and *word2vec*: given $\vec{x}$, $\vec{y}$ the vector representations of an entity and its occupation and given $\vec{c}$ the relationship between those concepts, it has been shown that $\vec{x} + \vec{c} \approx \vec{y}$ We therefore compute the embedding vector $\vec{a}$ of A and then identify set B of candidate entities for B, using the methods described before. We denote the embedding of the modifier C with $\vec{c}$.

The projection-based method relies on a different assumption. We project both $\vec{a}$ and $\vec{b}$ to the hyperplane perpendicular to $\vec{c}$ using

$$proj_{\vec{c}}(\vec{x}) = \vec{x} - \frac{\vec{x} \circ \vec{c}}{\vec{c} \circ \vec{c}} \vec{c},$$

where $\circ$ denotes the inner product.

We then extract a set of k candidates $\vec{b} \in B$ that maximise the cosine similarity between $proj_c(\vec{b})$ and $proj_c(\vec{a})$. Formally, we compute the similarity as $sim(\vec{a}, \vec{b}) = \frac{\vec{a} \circ \vec{b}}{|\vec{a}||\vec{b}|}$. Intuitively, this similarity function allows us to obtain a set of candidates for B whose representation is supposed to be similar to the one of A, after we "projecting away" the characteristics pertaining to C. To increase the creativity of a VA we propose to further rank the top-k results by using their $L_1$ norm. The intuition is that among all the candidates identified by sim, the ones that have a greater distance from the origin of the vector space are the most "extremal" ones.

```
PREFIX wdt: <http://www.wikidata.org/prop/direct/>
PREFIX wd: <http://www.wikidata.org/entity/>
SELECT DISTINCT ?item ?itemLabel ?occupation ?sitelinks WHERE {
         ?item wdt:P31 wd:Q15632617; #fictional characters
              wdt:P106 ?occupation;
              wikibase:sitelinks ?sitelinks .
         FILTER(30 < ?sitelinks).
         SERVICE wikibase:label {bd:serviceParam
                   wikibase:language   "[AUTO_LANGUAGE],en".}}
```

Listing 1: SPARQL Query to extract famous fictional characters

```
PREFIX wdt: <http://www.wikidata.org/prop/direct/>
PREFIX wd: <http://www.wikidata.org/entity/>
SELECT DISTINCT ?item ?itemLabel ?occupation ?sitelinks WHERE {
            ?item wdt:P31 wd:Q5;
                  wdt:P106 ?occupation
                  wikibase:sitelinks ?sitelinks;
            {?item wdt:P641 ?sport.} #people who do sports
    UNION   {?item wdt:P106 wd:Q82594.} #computer scientists
    UNION   {?item wdt:P106 wd:Q170790.} #mathematicians
    UNION   {?item wdt:P106 wd:Q82955.} #politicians
    UNION   {?item wdt:P106 wd:Q131524.} #enterpreneurs
    UNION   {?item wdt:P1303 ?instrument.} #people who play an instrument
    UNION   {?item wdt:P136 ?genre.} #people who have a genre
            FILTER(70 <= ?sitelinks).
            SERVICE wikibase:label{ bd:serviceParam
                        wikibase:language  "[AUTO_LANGUAGE],en".  }}
```

Listing 2: SPARQL Query to extract famous real-world individuals

At the end of our pipeline, the components of our Vossian Antonomasias are obtained, consisting of the source, target and context. These are then transformed into natural language in form of an assertion that connects the given elements using the following template: {A} {verb} the {B} of {C}. The verb is adapted according to the fact that the individual corresponding to the source has an entry for a date of death as a property in Wikidata. If this is not is the case, the verb becomes "is", otherwise "was" is inserted.

### 8.6.1   Purely LMM-based Baseline via ChatGPT

Finally, we will use ChatGPT [12] as a baseline model to compare the performance with a large transformer-based language model. After several attempts, we found the best combination of prompts to obtain the best possible results that meet the criteria of the Vossian Antonomasias:

- The target should not share characteristics with the modifier.

- The source and the target should share at least one salient characteristic.

- The source and the target should be popular enough to draw the analogy in the context of the modifier.

Our attempts at arriving at a good prompt template to generate a selection of VAs of a given target *A* via ChatGPT (together with an explanation for the particular choice) resulted in the following:

Provide 10 Vossian Antonomasias for *<Name of A>*, where she
is equated with another person. Each of the phrases should
have the structure "*<Name of A>* is the [person name] of
[profession]", where [profession] must not characterize
[person name]. Provide a very short justification for each
example.

For targets A known to identify as male, she would be replaced by he.

## 8.7    Evaluation

We want to propose an evaluation process that encompasses the results produced
by the knowledge graph embedding-based method, lexico-semantic embedding-
based method and ChatGPT. In this way, we can evaluate the results produced
by our pipeline using a mixed-method approach of symbolic representation and
embedding-based models in comparison to the output of ChatGPT as a repre-
sentation of a large language model on a broader level. At the same time we
compare between the knowledge graph embeddings and the lexico-semantic em-
beddings on a narrower level. We hope that by using the approach of knowledge-
based selection of the source, target and modifier, we reach a higher quality of
output in comparison to then lexico-semantic embedding method and large lan-
guage models like ChatGPT.

Table 8.1: Examples generated by three different methods

| Knowledge Graph Embeddings | Word Embeddings | ChatGPT (LLM) |
| --- | --- | --- |
| Alan Turing was the John Stuart Mill of computer scientists | Alan Turing was the Michael Keaton of computer scientists | Alan Turing is the Da Vinci of computer science. |
| Angela Merkel is the Warren Buffett of politicians | Angela Merkel is the Vincent van Gogh of politicians | Angela Merkel is the Mandela of diplomatic leadership. |
| Freddie Mercury was the Heidi Klum of singer-songwriters | Freddie Mercury was the Usain Bolt of singer-songwriters | Freddy Mercury is the Chaplin of stage performance. |

We chose three distinguished individuals to be the source for comparing
different approaches for the selection of a target that meets the criteria described
in section 8.3 and the selection of a modifier that is appropriate for the source.

The selection of the examples from Table 1 show that ChatGPT is not always
able to pick a source and a target from two separate domains, which can be seen
in the sentence "Angela Merkel is the Mandela of diplomatic leadership". The
same accounts for the example with Freddy Mercury as a source. The approach
using knowledge graph embeddings produces sentences that meet the criteria of distinguishing between the source and the target in the context of the modifier. The word embeddings also produced satisfying results that meet the mentioned criteria.

Due to the highly diverse nature of VAs, the quality of the output should be evaluated by human raters in a follow-up study. At first, we would provide a sample with 10 sentences for each method. To guarantee comparable results, we will randomly sample a certain number of sources that serve as the basis of the generation. To ensure comparability, we want to generate a sufficient amount of sentences to guarantee an equal distribution between the domains mentioned in chapter 8.7.The generated output can then be rated in form of a Likert Scale ranging from 1 to 5. Apart from the quantitative evaluation of the output's quality, the knowledge about the source and the target will be inquired in form of questions. This ensures a distinction between a negative rating caused by ignorance of the output's components and the lack of a proper connection between the source, the target and the modifier.

## 8.8   Discussion and Conclusions

We looked at generating Vossian Antononmasias by using embeddings and LLM as a way to characterize the creativity of AI. Our approach has resulted in creative examples of VAs, which proves that both method are suitable to solve such tasks. The lack of a clear definition of creativity prevents a quantitative evaluation of the results. We conducted a manual qualitative analysis on the results which highlighted several different weaknesses. The information bias that is inherent to Wikidata results in VAs that are mostly focused on western culture. While this might be tampered through a penalisation of such entities, we argue that such an approach would only partially solve such issue. A different approach, which takes into account the semantic representation of each entity, can help overcoming such issues, making the creation process transparent and explainable. For future work, we envision a qualitative evaluation of the VAs, to get a better understanding to which degree the described methods work and we plan to test more recent embedding methods, which have proven to be effective in other domains [7, 87].

# Chapter 9

# Improving Tacit Knowledge Acquisition

ALEKSANDRA SIMIĆ, BEATRIZ OLARTE MARTINEZ,
BOHUI ZHANG, INÊS KOCH, PHILIPP HANISCH,
SOULAKSHMEE D. NAGOWAH, HARALD SACK

## 9.1 Semantic Web for Creative AI

The underlying process of knowledge extraction from (human) experts involves language to communicate this knowledge. However, not everything we know has been written down, has been said, or is available as training data for LLMs. Moreover, "we can know more than we can tell" (Polanyi's Paradox). Considering the recent rise of popularity of Large Language Models (LLMs) and, in particular, ChatGPT, the question arises whether LLMs are capable of achieving. In their very nature, LLMs rely on language and, thereby, our capacity to formulate and write down knowledge. However, there are several areas I- ing creative tasks, e.g., writing funny texts, social skills, e.g., sarcasm detection, and physical skills, e.g., crocheting, which are inherently elusive. Even though we – as humans – have an implicit understanding of these activities, we struggle to formalize what we are doing. This idea that "we can know more than we can tell" was an insight by Polanyi who coined the term *tacit knowledge* [134]. We approach the concept of tacit knowledge with different approaches of the Semantic Web in mind. Additionally, we are discussing tacit knowledge from the perspective of LLM. Understanding what – and what not – we are able to formalize, can help to identify limitations as well as potentials of different formalism. Tasks related to tacit knowledge might benefit from the combination of different methods, e.g., combining LLMs with ontologies, or a new approach.

## 9.2 Introduction

The change from the information society to the knowledge society was already reported in 1992 [176]. More than thirty years later, we still rely on a master to transfer her/his knowledge of performing some manual skills to an apprentice as we struggle to verbalize and formalize how to execute the skill. On the other hand, the support of epistemological modeling frameworks such as TMDA [117] or CommonKADS [148] has helped many organizations in the formalization of explicit knowledge. Thus, there is a distinction between explicit and tacit knowledge, and the question arises whether the recent advances in the open access to Artificial Intelligent (AI) chatbots, e.g., the success of ChatGPT, can help to formalize (former) tacit knowledge, hard to put in words. Considering the lack of manual labor [61] and the need for digitization of knowledge in Industry 4.0 [164], this task of knowledge formalization — an old subject from AI – is more important than ever. Indeed, Large Languages Models (LLMs), such as ChatGPT and BERT, have disrupted society, yet they seem to be a threat to many workers. To see if modern AI approaches such as LLMs support verbalization and formalization of tacit knowledge, we formulate the following research questions:

**RQ1** – How can we identify tacit knowledge and its categories?

**RQ2** – Can tacit knowledge be formalized?

**RQ3** – Can the usage of ChatGPT improve the verbalization of tacit knowledge?

This paper follows the following structure: Section 3 provides the definition of tacit knowledge. Section 4 presents related works where ontological models are used to represent tacit knowledge and the investigation of tacit knowledge in LLMs. Section 5 describes the experiments performed to evaluate the current situation of tacit knowledge and LLMs and their results. In Section 6, after the previous results, we propose an approach towards tacit knowledge formalization. Finally, Section 7 summarizes the findings.

## 9.3 Tacit knowledge

Knowledge can be divided into several categories. Tacit knowledge is a proper subset of implicit knowledge (i.e. hard to verbalize and not consciously known). To explain the meaning of tacit knowledge, we firstly introduce the contrasting type of knowledge, explicit knowledge. Explicit knowledge is formalized and can be conveyed using a set of procedures. Examples include mathematical equations, the set of rules for assembling furniture, or a manual for using a specific electronic device. On the other hand, Mohameda, Leeb, and Salimc[113] highlight that tacit knowledge is learned through social experience and it is hard to formalize [113]. They also suggest methods for changing implicit information into explicit knowledge, and vice versa.

Within this paper, we understand tacit knowledge as *knowledge that is hard to formalize, articulate, and transfer; it is implicit, typically acquired through personal experience, learned through sensory interaction, and can be context dependent, e.g., cultural background.* We divide tacit knowledge into two groups:

- Tacit knowledge related to knowledge about skills acquired through **social interaction**. An example for this type is the way people communicate (both orally and by using mimics, gestures and prosody ) with each other and understand certain characteristics of a person such as trustworthyness, honesty, or sense of humor. Additionally, people have an inherent awareness of how to detect sarcasm (i.e. according to the [1], "the use of remarks that clearly mean the opposite of what they say, made in order to hurt someone's feelings or to criticize something in a humorous way") by experiencing different situations and learning from them.

- Tacit knowledge that is related to performing specific **physical activities**, such as riding a bike, pottering, playing an instrument or sports. Very often we call a person that has specific ability as talented, which is tightly related with the ability to perform a certain set of skills.

## 9.4    Related  Work

Researchers have investigated on the use of ontological models to represent and transfer tacit knowledge. There have been several ontological models proposed to represent and transfer tacit knowledge. With the emergence of LLMs, there are also works that investigate the ability of LLMs to understand tacit knowledge through the aspect of humor.

### 9.4.1    Tacit Knowledge in Ontological Models

Mohameda, Leeb, and Salimc have proposed an ontology-based knowledge model for software experience management. The authors represent organizational knowledge in terms of individual knowledge and group knowledge and further splits the organizational knowledge into tacit and explicit. They refer to tacit knowledge as perceptions, viewpoints, beliefs, and know-how. Tacit knowledge is further categorized as cognitive knowledge, such as viewpoints, mental models, conceptions, and technical knowledge with tacit elements, such as know-how and skills. The knowledge model entitled K-Assets consists of four ontologies, namely (i) competence ontology; (ii) information ontology; (iii) type ontology, and (iv) history ontology. The information ontology models the lessons learnt [113].

Rao and Nayak came up with an Enterprise Ontology model and patterns that are important in externalizing the tacit knowledge of socio-technical enterprises. The authors emphasize the fact that enterprise knowledge is highly tacit and underutilized. The methods used for capturing enterprise knowledge involve questionnaires and interviewing, storytelling, concept maps, wikis and social media, joint application design, and enterprise blogs. Two main actions for externalizing knowledge consisted firstly of interactions between the knowledge seekers and experts and, secondly, the production of explicit knowledge [140].

To bridge the gaps in current health information systems, Fareedi, Ghazawneh, and Se highlights the various approaches to knowledge acquisition in terms of tacit and explicit knowledge management that can be useful for capturing, coding, and communicating in the medical unit. They propose an ontology-based approach to address tacit knowledge acquisition issues and information flow problems. To collect data, two separate workshops were held, followed by a personal interview with the head nurse [64].

Chergui, Zidat, and Marir have proposed an ontological model to represent tacit knowledge in an organization. Interviews and self-confrontation techniques were used to identify tacit elements such as know-how a person possesses. The authors use the term 'know-how' to represent knowledge-in-act and theorem-in-act. A theorem-in-act refers to any proposition which is held true—whether rightly or wrongly—by a person in a given situation, and the concepts-in-act are what is considered pertinent by a person (what must be taken into account to succeed). Rule-based reasoning is used to identify the tacit knowledge and provide further explication [48].

Nordsieck, Heider, Winschel, and Hahner presents an approach to gather semantic knowledge of the manufacturing process and build a corresponding knowledge graph, which includes the tacit knowledge of experts during the manufacturing process. In this study, they conducted two different experiments considering decentralized knowledge extraction. To evaluate the proposed methodology, the authors apply their approach to plastics-based Fused-Deposition-Modelling, an additive manufacturing technique, and adjust the formalization accordingly [125].

### 9.4.2 Tacit Knowledge in LLMs

With the prosperity of LLMs, such as GPT-4 [129], works that try to understand and capture the tacit knowledge inside LLMs have emerged. A sense of humor, as an example of tacit knowledge, has been scrutinized for LLMs by prompting ChatGPT by asking it to tell jokes. It is found that ChatGPT is mostly repeating 25 jokes in 1008 examples. After prompting for explanations, it has also been shown that ChatGPT's ability to explain these jokes accurately and reasonably [86]. Chen et al. evaluate the understanding of humor of pre-training language models through four representative tasks, including humor recognition, humor type classification, humor level classification, and punchline detection [47].

## 9.5 Experiments and Results

The objectives of the evaluation are (i) to investigate how LLMs can handle tasks involving tacit knowledge and (ii) to show that written or verbal instructions without further input struggles with teaching physical skills involving tacit knowledge. Due to the human dependency of sarcasm and manual tasks a specific scientific methodology was developed, by experimental evaluation, designing a set of three experiments and their subsequent evaluation.

As an instance of the first objective, we evaluate how good ChatGPT is at detecting sarcasm in news headlines. Regarding the second objective, we perform two qualitative experiments on teaching basics of crochet and a dancing performance.

### 9.5.1 Sarcasm Detection

We evaluate sarcasm detection by ChatGPT on the data set by Misra and Arora [112], which the authors used to evaluate sarcasm detection by other systems. The data set contains news headlines which are either taken from *HuffPost* [1], who publishes serious headlines, and *TheOnion*[2], who purposely publishes sarcastic headlines. Taking the chat-like interface of ChatGPT into account, we use the first 50 sarcastic and 50 non-sarcastic headlines of the data set, and we shuffle the resulting set of hundred headlines randomly. We prompt ChatGPT by the task "Decide whether the following news headlines are sarcastic or not:", followed by the list of headlines. We evaluate the answer by ChatGPT against the classification given by the data set.

The results are shown in Figure 9.1: out of the 50 sarcastic headlines, Chat-GPT classified 28 correctly as sarcastic and 22 falsely as non-sarcastic, and it classified 43 out of the 50 non-sarcastic headlines correctly. All in all, ChatGPT reached an accuracy of 0.71, a recall of 0.56, and a precision of 0.80 on the (reduced) data set.

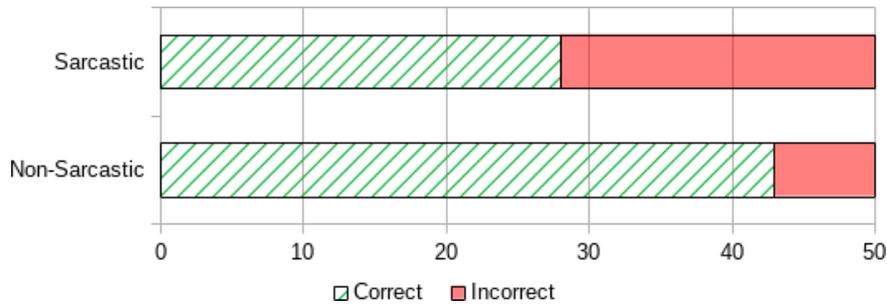

Figure 9.1: Classification of news headlines by ChatGPT; ChatGPT classified 28 out of 50 sarcastic and 43 out of 50 non-sarcastic news headlines correctly

In the experiments by Misra and Arora, the authors measure the performance of the model by Amir, Wallace, Lyu, Carvalho, and Silva [9] and their model: the former reached an accuracy of 0.8488, while the latter achieved an

---

accuracy of 0.897. Note that their evaluation takes the whole data set for training and evaluation into account. We see that ChatGPT can detect sarcasm, but is out-performed by models fine-tuned for sarcasm detection.

For a qualitative discussion, we consider a sarcastic headline that ChatGPT considered non-sarcastic: "NASA Now Almost Positive Mars Is Rocky". When specifically asked about this example, ChatGPT could relate that this is a straightforward and obvious observation and based on NASA's findings. Yet, it fails to realize that the sarcasm arises particularly from the obviousness of the statement. This highlights that sarcasm goes beyond easily expressible knowledge and requires tacit knowledge.

Table 9.1: Output of Crochet and Dancing experiments

| Crochet Experiment | | |
|---|---|---|
| | ChatGPT | Video |
| P1 | YES | YES |
| P2 | NO | NO |

| Dancing Experiment | | |
|---|---|---|
| | Verbal | Video |
| P1 | YES | YES |
| P2 | NO | YES |

## 9.5.2 Teaching Crocheting and Dancing

For both experiments about teaching tasks involving tacit knowledge, we invited two participants P1 (female) and P2 (male), both participants had no contact with the task for at least thirty years or had no previous contact at all. Their education level was the highest level of education possible, although in other domains. The results of these experiments are summarized in Table 9.1.

The first experiment was about learning how to **crochet**, a craft that involves creating fabric by interlocking loops of yarn or thread with a hooked needle. The first phase of this experiment is performing crochet yarn following ChatGPT instructions generated by the prompt: "*Can you explain to me how to crochet yarn?* ". In the second phase, participants are allowed to follow a video tutorial and do the same task again. After the experiments, we asked the participants to share their feelings and describe the knowledge learned from both phases of the experiments.

P1 was capable to perform the task in both cases. He/She experienced issues in starting the task by just reading the instructions provided by ChatGPT and could not detach the knitted piece following its instructions. The comments from the participant related to the experience with ChatGPT instructions are that they were very efficient in relation to the 'whats' but had many gaps about the 'hows', e.g., how to hold the hook properly. The participant learnt and intuitively inferred the gaps, but she felt that some information was missing. The immediate crochet didn't happen. P2 emphasized that     pecial instructions, such as backward or forward movements, were hard to understand by following the text given by ChatGPT. Both participants have not made any significant improvement by watching the video with instructions.

The second task that the participants had to perform was related to learning a **dance**.

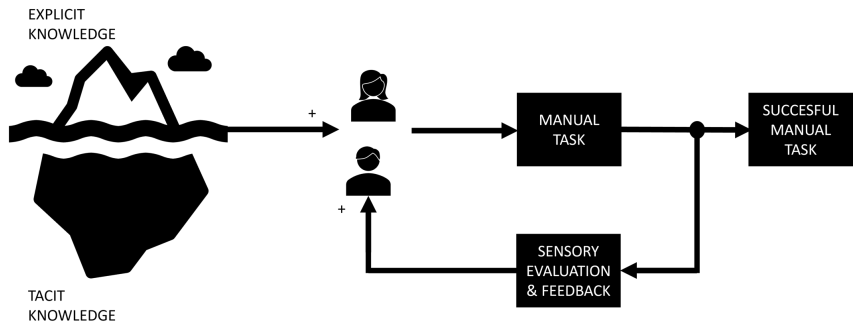

Figure 9.2: Feedback Loop to improve tacit knowledge acquisition

Both participants first received only verbal instructions related to the sequence of steps they had to perform during the dance. In this first stage, the participants did not have access to any rhythm or visual gestures that could help in understanding the proposed dance steps. The verbal instructions were biased, as they were given by a person who is not a professional dancer but who has knowledge about the steps. After receiving the directions, the participants listened to the music referring to the dance steps and were told to dance according to the instructions given.

In the second part of the task, each of the participants watched a video where they were shown the instructions for the same steps asked earlier. They were then asked to repeat the dance again with the background music.

The first participant (P1), as she was already an experienced dancer, responded to the oral instructions in a manner most similar to the original choreography. However, she performed much better when given both visual and oral instructions. The second participant (P2), who had no experience with dance, performed much better after watching the video.

According to the video recording and the self-description of the two participants, it could be observed that both of them can generate personal visual imagination based on the verbal dance instructions with different understandings of the rhythm (i.e., duration time of each movement) and the spatial aspects (i.e., the exact position of the hands and feet), as reported previously in [63]. In the present experiments, it was possible to observe that participants can make the basic movements according to the description. By enriching the visual information through video tutorials, it was possible to prove that their actions gradually become similar, which, in their words, was "intuitive".

## 9.6 Improving tacit knowledge acquisition

As shown in the previous sections, the approach adopted in this study relies on the fact that sensory systems and face-to-face human interactions [97] are required to succeed in the effective transfer of tacit knowledge. Therefore, a feedback loop, as depicted in Figure 9.2, analogous to other sciences, i.e. Systems Theory, is needed. The initial input to the process consists of explicit knowledge, as e.g. provided via LLMs or ontologies, which is used by the user for the manual task implementation. While the user performs this task feedback can be received either explicitly via language or tacit via sensory stimuli. The acquired feedback is leveraged to improve the knowledge about the manual task. This type of feedback is just available in hybrid approaches [49] or in full immersive platforms, as e.g., metaverse [62] providing enough support and acknowledgement of the surrounding environment to produce a real image of the objects and task in the user related supporting the learning process.

## 9.7 Discussion and Conclusions

Tacit knowledge is a valuable and often untapped resource within organizations. Our report performed three experiments, proposed insights for using LLMs, and accompanied knowledge in the modality of video. Although LLMs have the ability to capture and articulate tacit knowledge, this report suggests adopting hybrids or full immersive virtual learning approaches to improve tacit knowledge acquisition. By recognizing, capturing, and leveraging tacit knowledge, organizations can unlock hidden expertise, drive innovation, and foster continuous learning. Embrace the power of tacit knowledge to gain a competitive edge and thrive in a rapidly evolving world. Future steps in this field will use augmented reality and metaverse experiences to analyze user learning improvements.

# Chapter 10

# Exploring the Role of Generative AI in Constructing Knowledge Graphs for Drug Indications with Medical Context

REHAM ALHARBI, UMAIR AHMED, DANIIL DOBRIY, WERONIKA ŁAJEWSKA, LAURA MENOTTI, MOHAMMAD JAVAD SAEEDIZADE, MICHEL DUMONTIER

## 10.1  Research Question

The overall aim of the research is to develop and evaluate a framework to construct a knowledge graph from natural language sentences using LLM and entity recognition web services. The framework is applied to the problem of constructing a knowledge graph for drug indications, including their medical context, as part of creating a more complete knowledge graph of a drug's therapeutic intent.

We aim to answer the following research questions:

A. To what extent does the addition of more instruction to the LLM yield a more accurate/complete extraction of the context?

B. How many drug indications include a medical context?

C. What is the quality of the generated knowledge graph?

## 10.2 Semantic Web and Creative AI

When Semantic Web technologies intertwine with Creative AI, they utilise the richness and interconnectedness of medical data in a remarkable way. In the context of our research, the convergence of these technologies enables AI to construct a medical knowledge graph from natural language sentences, harnessing large language models (LLMs). This is particularly applicable to the complex task of creating a knowledge graph for drug indications, accounting for their medical context and/or therapeutic intent. The machine-readable data from the Semantic Web not only empowers AI to derive insights but also facilitates the construction of this knowledge graph. Through Creative AI's innovative capabilities, groundbreaking solutions to complex medical challenges can be realised. This collaboration results in a synergetic output that has the potential to revolutionise precision medicine, refine diagnostic accuracy, and considerably enhance healthcare delivery. Furthermore, it allows us to explore critical research questions, such as the impact of additional instruction to the LLM on extraction accuracy, the extent of medical context inclusion in drug indications, and the quality of the generated knowledge graph. This blend of Semantic Web technologies and Creative AI is a testament to the compelling potential of integrated tech in the medical realm.

## 10.3 Introduction

Drug indications are regulatory-approved uses for a drug. For example, paracetamol is indicated to treat headaches, muscle pain, and fever. The indication of a drug, along with how it should not be used (contra-indications), undesirable effects (adverse effects), and prescribing guidance for physicians is contained within drug product labels and is referred here to as the *therapeutic intent*. Drug product labels are package inserts whose contents are subject to approval by regulatory agencies such as the US Food and Drug Administration (FDA) or the European Medicines Agency (EMA). Crucially, the use of a drug may also be constrained to a particular medical context for which it has been demonstrated to be safe and effective. Valid medical contexts include the underlying medical illness for the target condition (e.g., *Quetiapine tablets are indicated for the acute treatment of manic episodes associated with bipolar I disorder* ), the age group (e.g., *EMGALITY is indicated for the treatment of episodic cluster headache in adults*), other conditions (e.g., *Clozapine is indicated for reducing the risk of recurrent suicidal behaviour in patients with schizophrenia*), and other drugs that should be pre- or co-administered (e.g., *Clonidine hydrochloride injection is indicated in combination with opiates for the treatment of severe pain in cancer patients*).

Machine-readable representations of drug indications are key in computational drug discovery and clinical decision-making [121]. However, while databases have been created to store drug indications in a machine-readable manner (e.g. DrugCentral [19]), these do not contain their medical context, and hence to not correctly cover the therapeutic intent. The lack of medical context will necessarily limit the development of accurate methods to predict new drug uses [121] or make treatment recommendations. As a case example, the Oncology Expert Advisor, a system derived from IBM Watson, was pulled from market after making unsafe suggestions relating to cancer treatment.[1]

Knowledge graphs (KGs) serve as a natural and intuitive way to store, query, and explore structured knowledge. However, building high-quality knowledge graphs requires substantial manual effort. Large Language Models (LLMs) are promising technologies for natural language understanding, and in particular, for constructing knowledge graphs. Recent unpublished work suggests that carefully engineered prompts have the potential to extract entities and their relations in a structured manner [162, 38]. However, there is no method that we are aware of that correctly extracts the medical context for drug indications, and hence the therapeutic intent.

**Contributions** In this work, we propose a novel approach to use a LLM that extracts drugs, their indications, and the associated medical context to a target Knowledge Graph. The main contributions of the research are as follows:

A. the development of a LLM based framework to extract a triple and its context, perform entity recognition, and produce a valid RDF graph

B. the use of the framework to extract drug indications and their medical context from sentences in natural language

C. an evaluation of prompts that vary in their specificity

D. an estimate of the scale of the problem of recognising and representing medical context

## 10.4  Related Work

The representation of therapeutic intent should define relationships between diseases and symptoms [121]. Additionally, various aspects that need to be taken into account while deciding on a treatment, such as disease progression, past treatments, mutations, or preexisting medical conditions provide a medical context that should be taken into account while creating accurate, precise, and compute-accessible representations of therapeutic intents. Drug indications may significantly vary depending on the context. The broad range of aspects to be contained in the representation makes the logic of therapeutic intent particularly complex. Capturing and structuring therapeutic intents for computational reuse is the first step toward precise computer-aided medicine. Therefore, an accurate, precise, and contextualized representation of the medical context is crucial. Otherwise, we risk that the algorithms relying solely on the diagnosis will lead to unjustified and potentially harmful results.

---

[1]https://www.statnews.com/2018/07/25/ibm-watson-recommended-unsafe-incorrect-treatments/

The task of synthesizing information about the therapeutic intent has been addressed in several works. MEDI [175] applies NLP and ontology relationships to extract indications for prescribable, single-ingredient medication concepts and all ingredient concepts. Drug-bank [178] combines detailed drug (i.e. chemical) data with comprehensive drug target (i.e. protein) information. DrugCentral [19] provides up-to-date drug information such as pharmacokinetic properties and a sex-based separation of side effects. All of these resources comprise structured information about the therapeutic usage of drugs. However, the medical context information is not part of them. Similarly, a number of computational methods have emerged to structure and utilize text-extracted drug indications. For instance, the PREDICT[19] algorithm employs drug and disease similarities to anticipate potential drug indications, validating its predictions via clinical trials and tissue-specific expression data of drug targets. [122] focus on automatically extracting and integrating drug indication information from multiple health resources such as DailyMed and MeSH Scope notes. By using state-of-the-art NLP methods, the paper contributes to the systematic construction of a disease/drug information repository and presents methods for gathering reliable drug indication information. However, to the best of our knowledge, none of the methods proposed for synthesizing drug indication information take into account medical context.

InContext is a human curated set of drug indications and medical context from the annotation of HTML pages from Dailymed, official provider of FDA label information. It contains annotations, including certain medical contexts, for 150 drugs related to neoplasms and cardiovascular diseases [114]. The corpus comprises both drug indications and contextual information that were manually annotated with the aid of Hypothesis.is [82] and the National Center for Biomedical Ontologies (NCBO) BioPortal annotator [25]. Each entity recognized by the annotator was assigned a corresponding MetaThesaurus/ontology term from the UMLS and/or Disease Ontology as an additional tag.

The NeuroDKG [115] is a Neuropharmaceutical Drugs Knowledge Graph, which contains drug indications and their medical context for a subset of drugs that target conditions of the nervous system. The database was created using Nanobench, a tool to create RDF nanopublications with auto-complete support for entity recognition.

## 10.5 Resources

### 10.5.1 Dailymed dataset

Dailymed is a comprehensive dataset, hosted by the National Institute of Health (NIH), that contains information about all the marketed drugs that have been approved by the U.S. Food and Drug Administration (FDA). It contains a wide range of information about drug names, drug classes, ingredients, indications, uses, dosages, contradictions, warnings, and precautions. The dataset are available as XML files through data dumps and an API.

For our experiments, we use the REST API to collect the following data:

A. The *Set Ids* and *Titles* of 145,601 drug labels[2]

B. The *Indications and Usages* sections for 4000 randomly sampled *Set Ids*[3]

## 10.5.2  InContext dataset

InContext comprises manually annotated concepts for 150 drugs. Each pharmaceutical formulation is identified by its Set Id, while the drug name, disease, and context, such as co-therapies and temporal aspects were identified through entity recognition. All concepts are mapped to UMLS Concept Unique Identifiers (CUIs), while diseases are also mapped to Disease Ontology Identifiers (DOIDs). InContext provides five different contexts:

A. Co-prescribed medication - drugs commonly prescribed together;

B. Co-therapies - procedures or therapies (not drug-related, i.e. radiotherapy) that should be applied in combination with the drug;

C. Co-morbidities - diseases or conditions that commonly occur together (with a target condition) in the same patients;

D. Genetics - genetic variants for a given disease;

E. Temporal aspects - information which explains at what life stage, disease stage, or treatment phase a drug should be administered.

## 10.5.3  NeuroDKG dataset

The NeuroDKG [115] dataset contains 2,397 triples, 460 entities, 13 properties, and 10 classes. It provides a wide range of relationships, presenting connections among various medical terminologies.

In our exploration of pharmacological data sources, we have integrated the Dailymed dataset along with the NeuroDKG and InContext datasets, presenting a more comprehensive and varied landscape of drug information. Dailymed, an invaluable dataset from the NIH, provides a wealth of details about FDA-approved drugs, complementing the focus areas of our other data sources. It brings together information about drug names, classes, ingredients, indications, uses, and much more, providing a rich context for our study. NeuroDKG, with its structured data on neuropharmaceutical drugs and relationships with medical terminologies, presents an intricate map of knowledge. On the other hand, InContext emphasizes anti-cancer and cardiovascular drugs, adding a layer of expert opinion on therapeutic indications and disease terms. With the inclusion

---

of the Dailymed dataset, our research now spans a broader spectrum of drug-related information, offering a more holistic and in-depth understanding of the pharmacological domain. Nanobench's contribution with a data model that encapsulates the medical context of a drug indication further strengthens our endeavour. This synergistic integration of varied datasets promises a richer and more nuanced exploration of drug data, paving the way for innovative healthcare research and solutions.

## 10.6 Proposed approach

We propose a framework to transform a textual description of drug indications with medical context into a knowledge graph using LLM-powered entity recognition along with identifier resolution (see Figure 10.1).

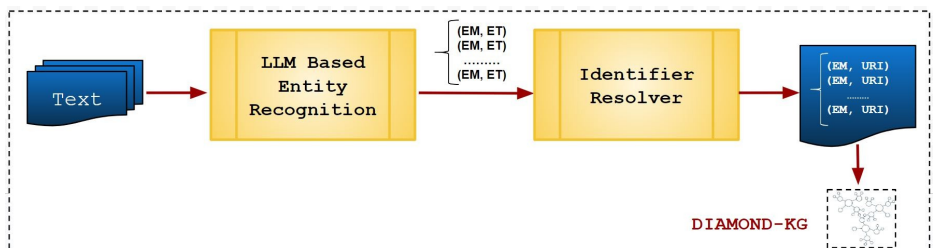

Figure 10.1: The DIAMOND-KG architecture

### 10.6.1 LLM-based entity recognition

We designed a set of prompts to guide a LLM to identify and extract entities of interest. Of particular interest is the extraction of contextual information beyond a simple statement consisting of a subject, verb and object. The prompts contain a *Paragraphs* section and a *Instructions* section. The prompts' instructions range from general instructions (Prompt 1) to specific (Prompt 3):

- Prompt 1: Triple Prompt; this prompt generates subject-verb-object triples from a given text in JSON format, such as *Convert subject-verb-object in the given text into a triple*, the corresponding prompt is shown in Figure 10.2.

- Prompt 2: Context Prompt; this is to identify the context from a given text and add this to the original triples such as *Convert subject-verb-object in the given text into triple, recognise conditional statements for the triple, and add them to the graph with the original triple*, the corresponding prompt is shown in Figure 10.3.

- Prompt 3: Medical Context Prompt; the given prompt parses a set of sentences and identifies medical context words or phrases within each sentence. The corresponding prompt is shown in Figure 10.4. This prompt contains a *Definition of context types* section within the *Instructions* section. The list of types of entities describing medical context for drug indications is presented in Table 10.1.

Table 10.1: Types of entities related to the medical context

| Medical Context | Description |
| --- | --- |
| age group | Refers to information related to the age group of individuals affected by a medical condition or receiving treatment. |
| co-morbidity | Refers to diseases or conditions that commonly occur together with a specific target condition in the same patients. |
| symptom | Refers to specific indications or symptoms associated with a medical condition. |
| co-therapy | Refers to procedures or therapies that should be used in combination with a particular treatment. |
| adjunct therapy | Refers to additional therapies or treatments used alongside the primary treatment. |
| past therapies | Refers to previous or historical treatments that a patient has undergone. |
| treatment duration | Refers to the duration or length of time for which treatment is recommended or prescribed. |
| conditional | Refers to statements or conditions that determine when the medication or treatment is appropriate to use. |
| co-prescribed medication | Refers to medications that are commonly prescribed together with a specific drug. |
| genetics | Refers to genetic factors or specific genetic strains associated with a disease. |
| temporal aspects | Refers to information related to the timing or stages of a disease or treatment, such as at what life stage or disease stage a drug should be administered. |

Each prompt is extended with additional specification to return a defined JSON-formatted object.

```
Paragraphs (separated by a full stop): {" ".join(newSentences)}

Instructions:

Inside each Paragraph, identify all triples and output a JSON dictionary with Paragraphs as keys
and lists of triples formatted as [SUBJECT, PREDICATE, VALUE] as values.
Only output the resulting JSON.
```

Figure 10.2: Prompt 1: Triples

```
Paragraphs (separated by " | "): {" | ".join(newSentences)}

Instructions:
Inside each Paragraph, identify all context words/phrases and output a JSON dictionary with
Paragraphs as keys and lists of {{[context type]: [corresponding words/phrases as a list of values
(more than one possible)]}} as values.
Other context types are self-explanatory.
Only output the resulting JSON.
```

Figure 10.3: Prompt 2: Free Context

## 10.6.2   Identifier Resolver

The values for each context type is then grounded to database/ontology identifiers using the NIH NCATS Translator SRI Name Resolution API [4]. The name

---

[4] https://name-resolution-sri.renci.org/docs

```
Paragraphs (separated by " | "): {" | ".join(newSentences)}
Instructions:
Inside each Paragraph, identify all context words/phrases (allowed context types: "age group",
"co-morbidity", "symptom", "co-therapy", "adjunct therapy", "past therapies", "treatment duration",
"conditional", "co-prescribed medication", "genetics", "temporal aspects") and output a JSON
dictionary with Paragraphs as keys and lists of {[[context type]: [corresponding words/phrases as a
list of values (more than one possible)]]} as values.
Definition of context types:
"conditional" - a statement about when the medication is appropriate to use
"target" - the condition (symptom or illness) that is intended to be treated
"co-prescribed medication"- drugs commonly prescribed together with the given drug (not therapeutic
procedures! -> for that use "co-therapy")
"co-therapy" - procedures or therapies that should be applied in combination with the drug (not
medications or substances! -> for that use "co-prescribed medication")
"co-morbidity" - diseases or conditions that commonly occur together (with a target condition) in
the same patients
"genetics"-  particular genetic strains of a disease
"temporal aspects"-  information which explains at what life stage, disease stage, or treatment
phase a drug should be administered
Other context types are self-explanatory.
Only output the resulting JSON.
```

Figure 10.4: Prompt 3: Medical Context

resolution service takes lexical strings and attempts to map them to identifiers (curies) from a vocabulary or ontology. The lookup is not exact, but includes partial matches. For each entity mention, we obtain a list of 5 results representing possible conceptual matches, of which the first is the preferred choice, and the remainder are ranked by next preferred resource.

### 10.6.3 RDF Graph Generation

The last component of the framework is the generation of an RDF graph from the JSON result containing the extracted and named entities and relations. The data are mapped to standard ontologies and vocabularies, such as RDF Statements or RDF Graphs, Nanopublications, and Micropublications. This component is under development and subject to change.

### 10.6.4 Implementation of the method

The implementation is in python 3 and is released with an MIT license[5] and published as a GitHub repository.[6]. We use OpenAI *gpt-4* model API as the LLM. The number of maximum requested tokens (with 1 token approximately corresponding to 4 chars of English text[7]) is set to 6.144 (current maximum value for the *gpt-4* model) to allow batch prompts containing many paragraphs as input. The batch size is heuristically set to 10 paragraphs per prompt for Prompt 1 and Prompt 3. For Prompt 2, the paragraphs are processed individually (1 paragraph per prompt) as batch processing for specifically this prompt has been found to lead to complex interactions in the prompt results due to context categories being conceptualised for the batch of sentences together. We use RDFLib 6.3.2 to generate an RDF graph.

---

[5]refer to https://opensource.org/licenses/MIT
[6]available at https://github.com/semantisch/diamond-kg
[7]For detailed token calculation refer to: https://help.openai.com/en/articles/4936856-what-are-tokens-and-how-to-count-them

Table 10.2: Mapping between predicates in the NeuroDKG dataset and the corresponding entity types related to the medical context defined within DIAMOND-KG.

| NeuroDKG predicate | DIAMOND-KG label |
|---|---|
| hasAgeGroup | age group |
| hasComorbidity | co-morbidity |
| hasSymptom | symptom |
| hasCurrentMedication | co-therapy |
| hasTherapy | adjunct therapy |
| treatmentDuration | treatment duration |

Table 10.3: Results statistics. For each prompt, we provide the number of processed sentences, the total number of extracted triples, and the average number of triples for each sentence.

| Prompt | Number of Sentences | Total Triples | Average Triples per sentence |
|---|---|---|---|
| NeuroDKG | 174 | 790 | 4.54 |
| Prompt 1 | 99 | 187 | 1.89 |
| Prompt 2 | 82 | 296 | 3.61 |
| Prompt 3 | 163 | 313 | 1.92 |

## 10.7  Evaluation and Results: Use case/Proof of concept - Experiments

### 10.7.1  Experimental Setup

To evaluate our system, we focus on the NeuroDKG dataset[8], which provides manually curated triples extracted at the sentence level for each drug indication. In the original dataset, each row corresponds to a triple extracted from a given drug label. Each row provides information about the subject, predicate, and object of the triple, and in some cases, we also have the corresponding ontology concept for the object and the sentences from which such triples were extracted. To provide a comparable ground truth, we discarded all triples that did not have a corresponding set of sentences. After this step, we had a dataset comprising 790 triples, with an average of 4.54 triples for each sentence. As far as context is concerned, 62% of sentences from NeuroDKG contain medical context. We selected some predicates from NeuroDKG that were mapped to the context information extracted by the DIAMOND-KG prototype (see Table 10.2). For each prompt, Table 10.3 shows the number of processed sentences, the number of triples that can be successfully extracted from sentences, and the average number of extracted triples per sentence. We also report the same information for NeuroDKG as a comparison.

---

[8]available at https://github.com/MaastrichtU-IDS/neuro_dkg

### 10.7.2    First Prompt Analysis

The first prompt generates triples from a given text in a JSON format, without providing any domain-specific information or insights about the type of text we are submitting. With this approach, we are able to extract 187 triples from 99 sentences, with an average of 1.89 triples per sentence which is the lowest among the different prompts. This can be attributed to the general nature of the instruction which may affect the system's performance.

### 10.7.3    Second Prompt Analysis

The second prompt tests the power of generative AI tools to extract context information from text, without providing any specific context label or information about the task domain. To this end, we analyze the overlap between the context information we defined in Section 10.6 and the one extracted by DIAMOND-KG by using the second prompt. As reported in Table 10.3 with prompt 2 we are able to extract 296 triples from 82 sentences, with an average of 3.61 triples per sentence. The LLM Based Entity Recognition module identifies 105 different predicates ranging from domain-specific entities such as "medication", "treatment", and "symptom" to more general entities like "demographics" or "Publication". We discover that most predicates are synonyms or they are in different cases. This situation can lead to inconsistency among the KG predicates and increases the possibility of duplicate information. To mitigate this issue, additional processing step will be needed between the output of the LLM Based Entity Recognition module and the construction of the KG, otherwise the prompt should include a list of pre-defined predicates.

### 10.7.4    Third Prompt Analysis

In the third prompt, we include domain-specific information and the set of context entities we plan to extract, as defined in Section 10.6. We extract 313 triples from 163 sentences, with an average of 1.91 extracted triples for each sentence. Limiting the list of possible predicates results in a decrease in the number of extracted triples, which may also lead to lost information. However, the third prompt identifies 75% sentences containing medical context, which means that we are able to extract context information for more sentences than the NeuroDKG datasets.

As reflected in Figure 10.5, most sentences were associated with two context entities: **'age group'** and **'conditional'**, with the former having **58** associations while the latter having **36**. The least amount of sentences were associated with **'past therapies'** and **'genetics'** with **2** and **0** associations, respectively. While other context entities have **14.86** as an average amount of sentences associated with them. Apart from the aforementioned 4 entities, it appears to have an even distribution.

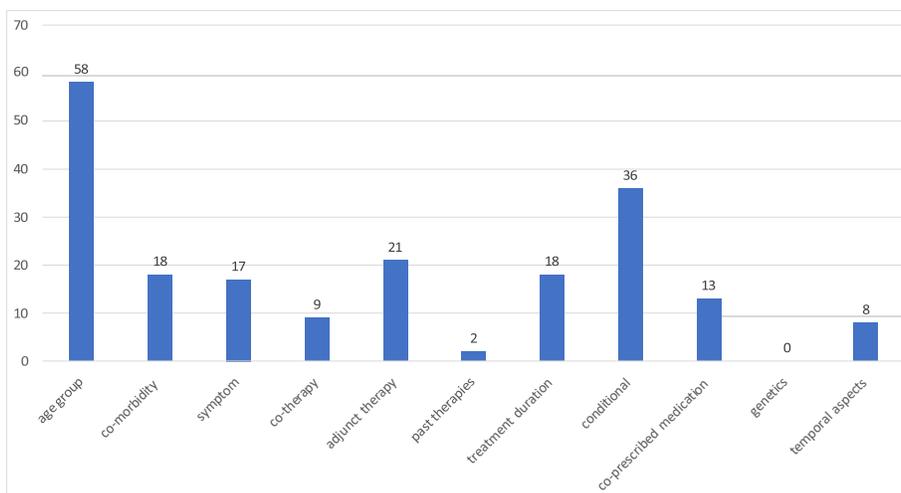

Figure 10.5: Context information distribution from prompt 3 results. For each context entity, we report the number of sentences for which such information has been extracted by the prompt.

## 10.8 Discussion and Conclusions

We propose a novel approach that leverages an LLM to extract drugs, their indications, and their associated medical context. To the best of our knowledge, this is the first such effort to do extract such content, and to do so with in an automated manner with LLMs. The prototype system uses the LLM to recognize entities, which are subsequently passed to a service to perform concept recognition, and that the final step creates a standards-compliant RDF knowledge graph that should be further augmented to meet the FAIR principles.

In relation to RQ1, we find that the framework generates similar extractions to the manually curated NeuroDKG dataset, but also finds a wider array of contexts. This raises the potential for this framework to accurately and systematically extract a wide variety of contextual information, and may have potential well beyond extracting drug indications and their medical context.

In relation to RQ2, the frequency of drug indications with a medical context was 62% in NeuroDGK, but our framework revealed that 75% of sentences had at least one context. This result suggests that i) a very large and significant proportion of drug indications do have a medical context, ii) that the framework is able to identify these to a greater extent than manual curation, and iii) the missing medical context is significant for applications focused on therapeutic intent.

In relation to RQ3, this question remains unanswered. Future work will need to investigate the quality of the generated knowledge graph.

# Acknowledgement

We would like to thank everyone who contributed to the organisation of ISWS, the students who are its soul and motivating engine, and the sponsors.

Please visit http://2023.semanticwebschool.org